# Moderately Supervised Learning: Definition, Framework and Generality

Yongquan Yang (remy_yang@foxmail.com)

## Abstract

Learning with supervision has achieved remarkable success in numerous artificial intelligence (AI) applications. In the current literature, by referring to the properties of the labels prepared for the training dataset, learning with supervision is categorized as supervised learning (SL) and weakly supervised learning (WSL). SL concerns the situation where the training dataset is assigned with ideal (complete, exact and accurate) labels, while WSL concerns the situation where the training dataset is assigned with non-ideal (incomplete, inexact or inaccurate) labels. However, various solutions for SL tasks under the era of deep learning have shown that the given labels are not always easy to learn, and the transformation from the given labels to easy-to-learn targets can significantly affect the performance of the final SL solutions. Without considering the properties of the transformation from the given labels to easy-to-learn targets, the definition of SL conceals some details that can be critical to building the appropriate solutions for specific SL tasks. Thus, for practitioners in various AI application fields, it is desirable to reveal these details systematically. This article attempts to achieve this goal by expanding the categorization of SL and investigating the sub-type that plays the central role in SL. More specifically, taking into consideration the properties of the transformation from the given labels to easy-to-learn targets, we firstly categorize SL into three narrower sub-types. Then we focus on the moderately supervised learning (MSL) sub-type that concerns the situation where the given labels are ideal, but due to the simplicity in annotation, careful designs are required to transform the given labels into easy-to-learn targets. From the perspectives of the definition, framework and generality, we conceptualize MSL to present a complete fundamental basis to systematically analyse MSL tasks. At meantime, revealing the relation between the conceptualization of MSL and the mathematicians' vision, this article as well establishes a tutorial for AI application practitioners to refer to viewing a problem to be solved from the mathematicians' vision.

# 1. Introduction

With the development of fundamental machine learning techniques, especially deep learning (LeCun et al., 2015), learning with supervision has achieved great success in various classification and regression tasks for artificial intelligence (AI) applications. Typically, a predictive machine learning model is learned from a training dataset that contains a number of training examples. For learning with supervision, the training examples usually consist of certain training events/entities and their corresponding labels. In classification, the labels indicate the classes corresponding to the associated training events/entities; in regression, the labels are real-value responses corresponding to the associated training events/entities.

In the current literature of learning with supervision, there are two main streams: supervised learning (SL) and weakly supervised learning (WSL) (Zhou, 2018). SL focuses on the situation where the training events/entities are assigned with ideal labels. The word 'ideal' here refers to that the labels assigned to the training events/entities are complete, exact and accurate. 'Complete' indicates that each training event/entity is assigned with a label. 'Exact' indicates that the label of each training event/entity is individually assigned. 'Accurate' indicates that the assigned label can accurately describe the ground-truth of the corresponding event/entity. In contrast, WSL focuses on the situation where the training events/entities are assigned with non-ideal labels. The word 'non-ideal' here refers to that the labels assigned to the training events/entities are incomplete, inexact or inaccurate. 'Incomplete' indicates that, only a proportion of training events/entities are assigned with labels. 'Inexact' indicates that several training events/entities can be simultaneously assigned with a same label. 'Inaccurate' indicates that the assigned label cannot accurately describe the ground-truth of the corresponding event/entity. More formal descriptions for SL and WSL, and their relations with this work are provided in Section 2.

The clear boundary between the descriptions of SL and WSL is the properties (completeness, exactness and accuracy) of the labels prepared for the training events/entities. However, in many real-world SL tasks under the era of deep learning, we cannot directly learn a predictive model that can effectively map the training events/entities to their correspondingly assigned labels. The main reason lies in the fact that the assigned labels are sometimes not easy to learn, though ideal (complete, exact and accurate) are they. This scenario appears due to the considerations of reducing the labour and difficulty in producing annotations for large amount of data under the era of deep learning. One must first transform the assigned labels into easy-to-learn targets for learning the predictive model of a SL solution. Existing solutions for various SL tasks have shown that the transformation from the given labels to the easy-to-learn targets can significantly affect the performance of the final SL solution (Law & Deng, 2020; Lin et al., 2017, 2020; Xie et al., 2018; Xue et al., 2019). By simply referring to the properties (completeness, exactness and accuracy) of the labels prepared for the training events/entities, and without considering the properties of the transformation from the given labels to the easy-to-learn targets, the definition of SL conceals some details that can be critical to building the appropriate solutions for certain specific SL tasks. Thus, for practitioners in various application fields, it is desirable to reveal these details systematically. This article attempts to achieve this goal by expanding the categorization of SL and investigating the central sub-type of SL.

Defining the properties of the transformation from the given labels to learnable targets for an SL task as 'carelessly designed' and 'carefully designed' two types, we further categorize SL into three narrower sub-types. The three sub-types include precisely supervised learning (PSL), moderately supervised learning (MSL), and precisely and moderately combined supervised learning (PMCSL). PSL concerns the situation where the given labels are precisely fine. In this situation, we can carelessly design a transformation to obtain the easy-to-learn targets from the given labels. In other words, the given labels can be viewed as easy-to-learn targets to a large extent. PSL is the most classic sub-type of SL, and typical tasks include simple task like image classification (Krizhevsky et al., 2012) and complicated task like image semantic segmentation (Ghosh et al., 2019). MSL concerns the situation where the given labels are ideal, but due to the simplicity in annotation of the given labels, careful designs are required to transform the given labels into easy-to-learn targets for the learning task. This situation is different from the classic PSL, since the given labels must be carefully transformed into easy-to-learn targets for the learning task, which would otherwise lead to poor performance. This situation is also different from WSL since the given labels are complete, exact and accurate. Typical MSL tasks include cell detection (CD) (Xie et al., 2018) and line segment detection (LSD) (Xue et al., 2019). PMCSL concerns the situation where the given labels contain both precisely fine and ideal but simple annotations. Usually, PMCSL consists of a few PSL and MSL sub-tasks. Typical PMCSL tasks include visual object detection (Zhao et al., 2019), facial expression recognition (Ranjan et al., 2019) and human pose identification (Cao et al., 2019). More detailed characteristics of PSL, MSL and PMCSL are provided in Section 3.

In the three narrower sub-types, MSL counts for the majority of SL, due to the fact that PSL only counts for a minor proportion of SL and MSL is an essential part of PMCSL which account for the major proportion of SL. As a result, MSL plays the central role in the field of SL. Although solutions have been intermittently proposed for different MSL tasks, currently, insufficient researches have been devoted to systematically analysing MSL. In this article, we present a complete fundamental basis to systematically analyse MSL, via conceptualizing MSL from the perspectives of the definition, framework and generality. Primarily, presenting the definition of MSL, we illustrate how MSL exists by viewing SL from the abstract to relatively concrete. Subsequently, presenting the framework of MSL based on the definition of MSL, we illustrate how MSL tasks should be generally addressed. Finally, presenting the generality of MSL based on the framework of MSL, we illustrate how generality exist among different MSL solutions by viewing different MSL solutions from the concrete to relatively abstract; that is, solutions for a wide variety of MSL tasks can be more abstractly unified into the presented framework of MSL. Particularly, the framework of MSL builds the bridge between the definition and the generality of MSL. More details of the conceptualization of MSL from the perspectives of the definition, framework and generality are provided in Section 4.

For an application practitioner who invents deep learning-based technologies to promote the development of AI applications, in addition to chasing the state-of-the-art result for a problem to be solved, viewing the problem from the mathematicians' vision is as well or even more critical to discover, evaluate and select appropriate solutions for the problem, especially when deep learning has been becoming increasingly standardized and reaching its limits in some specific AI applications. However, currently, most AI application practitioners primarily focus on chasing the state-of-the-art results for a problem to be solved, and few pay attention to viewing the problem from the mathematicians' vision. So, the question is how is the mathematicians' vision? There is a generalized answer to this question, which has been presented by the Chinese scientist Jingzhong Zhang (J. Zhang, 2016): "Mathematicians' vision is abstract. Those we think are different, they seem to be the same. Mathematicians' vision is precise. Those we think are the same, they seem to be very different. Mathematicians' vision is clear and sharp. They continue pursuing mathematical conclusions that we feel very satisfied with. Mathematicians' vision is dialectical. We think one is one and two is two, but they often focus on what is unchanging in the changing and what is changing in the unchanging."

What we see from this generalized answer is that can we gain intrinsic insight into the nature of a problem to be solved only when we look at the problem from the mathematicians' vision, which is at least both from the abstract to the concrete and from the concrete to the abstract. With this in minds, in this article, we show the definition of MSL exists when we view SL from the abstract to relatively concrete, while the generality of MSL exists among different solutions when we view them from the concrete to relatively abstract. As a result, intrinsically, the conceptualization of MSL presented in this article is the product of viewing a problem to be solved from the mathematicians' vision. More details about the relation between the conceptualization of MSL and the mathematicians' vision can be found in Section 5.

In a previous article (Yang & Zheng, 2020), the existence of MSL was discussed for the first time. Differently, in this article, in addition to more completely conceptualizing MSL, we focus more on how MSL is conceptualized (i.e., illustrating the underneath methodology of conceptualizing MSL). Conceptualizing MSL from the perspectives of definition, framework and generality, this article provides a complete fundamental basis to systematically analyse the situation, where the given labels are ideal, but due to the simplicity in annotation of the given labels, careful designs are required to transform the given labels into easy-to-learn targets. At meantime, revealing the intrinsic relation between the conceptualization of MSL and the mathematicians' vision, this article also establishes a tutorial for AI application practitioners to refer to viewing a problem to be solved from the mathematicians' vision. We hope this article will be helpful to realize that it is as well or even more critical to systematically analyse a problem to be solved, in addition to chasing the state-of-the-art result. In summary, the detailed contributions of this article are as follows:

- The conceptualization of MSL is more completely presented from the perspectives of definition, framework and generality.
- Viewing SL from the abstract to relatively concrete, how the definition of MSL was revealed is illustrated.
- The framework of MSL, which builds the bridge between the definition and the generality of MSL, provides the foundation to systematically analyse how MSL tasks should be generally addressed.
- Viewing different MSL solutions from the concrete to relatively abstract, how the generality of MSL was revealed is illustrated.
- The intrinsic relation between the conceptualization of MSL and the mathematicians' vision was revealed.
- The whole article establishes a tutorial of viewing a problem to be solved from the mathematicians' vision.

The rest of this article is structured as follows. In section 2, we discuss SL and WSL, and their relations to MSL. In section 3, we describe how SL is categorized into three narrower sub-types, discuss the relations of the three SL sub-types, and compare the SL sub-types with the WSL sub-types. In section 4, we present the definition, framework and generality of MSL, and illustrate how the definition and generality of MSL were revealed. In section 5, the relation between the conceptualization of MSL presented in this article and the mathematicians' vision is revealed to illustrate the underneath methodology of proposing the new concept of MSL. Finally, we discuss the whole article and point some possible future research directions for MSL in section 6.

## 2. Related Works

Formally, the task of learning with supervision is to learn a function $f: d \longmapsto g^*$ from a training dataset $\mathcal{T}$. Usually, $d$ denotes a set of events/entities, $g^*$ represents the given labels corresponding to $d$, $f$ is a function that can map $d$ into corresponding $g^*$, and the training dataset $\mathcal{T}$ consists of the events/entities $d$ and corresponding labels $g^*$. In the current literature, there are two main types: supervised learning (SL) and weakly supervised learning (WSL). Usually, these two types are distinguished according to the properties (completeness, exactness and accuracy) of the labels $g^*$ prepared for the events/entities $d$ in the training dataset $\mathcal{T}$. Both SL and WSL are related to the concept of MSL in this article, since the proposal of MSL is started from the clear boundary between SL and WSL.

### 2.1 Supervised learning

In the paradigm of SL, the predictive models are usually produced via learning with complete, exact and accurate labels. Specifically, the training dataset $\mathcal{T} = \{(d_1, g_1^*), \cdots, (d_N, g_N^*)\}$, where $N$ is the number of events/entities and each $d_n$ has a label $g_n^*$ that can ideally describe the ground-truth corresponding to $d_n$. Based on such carefully prepared training dataset $\mathcal{T}$, SL has been widely adopted to solve fundamental tasks such as image classification, visual object tracking, visual object detection, and image semantic segmentation (Yang et al., 2018, 2019, 2023; Yang, Lv, et al., 2020, 2021) and various other tasks (Li et al., 2021, 2022).

In this article, taking into consideration the properties of the transformation from the given labels to learnable targets in the SL paradigm, we categorize SL into three narrower sub-types and particularly conceptualize the MSL sub-type. More details about the three sub-types of SL will be discussed in section 3.

### 2.2 Weakly supervised learning

In the paradigm of WSL, the predictive models are usually produced via learning with incomplete, inexact or inaccurate labels (Zhou, 2018).

Learning with incomplete labels focuses on the situation, where only a small amount of ideally labelled data is given while abundant unlabelled data are available to train a predictive model. In this situation, the ideally labelled data are commonly insufficient to learn a predictive model that has good performance. Typical techniques for this situation include active learning (Settles, 2010) and semi-supervised learning (Zhu, 2008). Formally, the training dataset for this situation can be denoted as $\mathcal{T} = \{(d_1, g_1^*), \cdots, (d_J, g_J^*), d_{J+1}, \cdots, d_N\}$, where only a small portion $\{d_j | j \in \{1, \cdots, J\}\}$ of $d$ has labels.

Learning with inexact labels concerns the situation, where only coarsely labelled data is leveraged to learn a predictor. Multi-instance learning (Foulds & Frank, 2010) is the typical technique for this situation. Formally, the training dataset for this situation can be denoted as $\mathcal{T} = \{(d_1, g_1^*), \cdots, (d_N, g_N^*)\}$, where $d_n = \{d_{n,1}, \cdots, d_{n,M_n}\} \subseteq d$ is called a bag, $d_{n,m} \in d$ $(m \in \{1, \cdots, M_n\})$ is an instance, and $M_n$ is the number of instances included in $d_n$. For a two-class classification multi-instance learning task where $t^* = \{y, n\}$, $d_n$ is a positive bag, i.e., $t_n^* = y$, if there exists $d_{n,p}$ that is positive, while what is known is only $p \in \{1, \cdots, M_n\}$. The goal is to predict labels for unseen bags.

Learning with inaccurate labels focuses on the situation, where the data, for which the labels prepared may contain errors, is leveraged to train a reasonable predictive model. Learning with noisy labels (Frénay & Verleysen, 2014) is the typical technique for this situation. Formally, the training dataset for this situation can be denoted as $\mathcal{T} = \{(d_1, (g_1^* + \Delta_1)), \cdots, (d_N, (g_N^* + \Delta_N))\}$, where $(g_n^* + \Delta_n)$ is the given label corresponding to $d_n$, which consists of an accurate label $g_n^*$ and a label error $\Delta_n$. Possessing the advantages in efficiency and cost of the data labelling process, WSL has become popular in addressing various SL tasks that requires labour extensive annotations, such as some image analysis tasks (Yang, Li, et al., 2021; Yang, Yang, Chen, et al., 2020; Yang, Yang, Yuan, et al., 2020).

In this article, the presented SL sub-types are different from WSL sub-types, since the prepared labels for SL sub-types are ideal as they are in the paradigm of SL while the prepared labels for WSL sub-types are non-ideal. More details about the differences between the presented SL sub-types and WSL sub-types will be discussed in section 3.

## 3. Sub-types of Supervised Learning

The categorization of learning with supervision into SL and WSL in section 2 simply takes into consideration the properties (completeness, exactness and accuracy) of the labels prepared for the training dataset. However, in practice, we usually cannot directly learn a function $f: d \longmapsto g^*$ that can effectively map the events/entities $d$ into the corresponding labels $g^*$ for an SL task especially under the era of deep learning. Due to the considerations of reducing the labour and difficulty in producing annotations for large amount of data under the era of deep learning, the given labels $g^*$ are sometimes are not easy to learn. Usually, we must first build a transformation from given labels $g^*$ to easy-to-learn targets $t^*$, and learn a function $f: d \longmapsto t^*$ that can effectively map the given labels $g^*$ into the transformed easy-to-learn targets $t^*$. This scenario needs to be more systematically discussed.

In this section, by taking the properties of the target transformation from $g^*$ to $t^*$ into consideration, we expand SL into three narrower sub-types. Usually, a transformation from given labels to easy-to-learn targets for an SL task is coupled with a re-transformation from the predicted targets of the learnt function $f$ to the final predicted labels. Since a label re-transformation commonly consists of the reverse operations corresponding to its coupled target transformation, in this section, we assume that the properties of the label re-transformation remain the same as the properties of its coupled target transformation and give no additional discussions.

### 3.1 Properties of target transformation

We classify the transformations from given labels to easy-to-learn targets for SL tasks into 'carelessly designed' and 'carefully designed' two types. Intrinsically, we define that a target transformation is the 'carelessly designed' type if it is non-parameterized while a target transformation is 'carefully designed' type if it is parameterized, due to the fact that a non-parameterized target transformation simply requires careless designs while a parameterized target transformation must require careful designs. That a non-parameterized target transformation simply requires careless designs is because it can generate a type of the easy-to-learn targets that can be considered to be optimal. However, that a parameterized target transformation must require careful designs is because adjusting its parameters can generate various types of easy-to-learn targets from which the optimal type of easy-to-learn targets need to be found. To formally summarize the properties of a target transformation, we present Definition 1 as follows.

**Definition 1.** *For the given labels* $g^*$*, the easy-to-learn targets generated by a 'carelessly designed' target transformation are*

$$t^* = \left\{t_n^* \middle| t_n^* \approx g_n^*, n \in \{1, \cdots, N\}\right\},$$

*where* $\approx$ *signifies the non-parameterized target transformation of* $t_n^*$ *from* $g_n^*$ *and the type of targets* $t^*$ *can be considered as optimal; and the easy-to-learn targets generated by a 'carefully designed' target transformation are*

$$t^* = \left\{t_n^* \middle| t_n^* \lll g_n^*, n \in \{1, \cdots, N\}\right\},$$

*where* $\lll$ *signifies the parameterized target transformation of* $t_n^*$ *from* $g_n^*$ *and the optimal type of targets* $t^*$ *need to be found by adjusting the parameters of the target transformation.*

## 3.2 Narrower sub-types of SL

Taking into consideration the properties presented in Definition 1 for the target transformation, we further classify SL into three narrower sub-types: precisely supervised learning (PSL), moderately supervised learning (MSL), and precisely and moderately combined supervised learning (PMCSL).

### 3.2.1 Precisely supervised learning

PSL concerns the situation where the given labels $g^*$ in the training dataset have precisely fine ground-truths. In this situation, we can simply construct a non-parameterized target transformation with careless designs to obtain the easy-to-learn targets $t^*$ from $g^*$. Image classification (Krizhevsky et al., 2012) and image semantic segmentation (Ghosh et al., 2019) are two typical PSL problems.

In a $C$-class image classification task, the given ground-truth label for the class of an image can usually be transformed into an easy-to-learn target using a $C$-bit vector. In this vector, the bit corresponding to the given ground-truth label is set to 1, and the remaining bits are set to 0. Similarly, for a $C$-class image semantic segmentation task, each pixel point in the given ground-truth label for the semantic objects in an image can be transformed into a value at the same pixel point in the easy-to-learn target. The transformed value can be a one-hot vector in classification or real-value response in regression, corresponding to its predefined class in the given ground-truth label. We can note that the target transformation for these two PSL tasks are non-parameterized and can be simply built with careless designs. In other words, to some extent, the given labels $g^*$ can be directly viewed as easy-to-learn targets $t^*$ due to their precise fineness.

In summary, a concise illustration for PSL can be shown as Fig. 1. For the image classification task of PSL (top row in Fig. 1), the event/entity ($d$) is an image lattice, the label ($g^*$) corresponding to the image lattice $d$ is a predefined category, and the target ($t^*$) is transformed from the label $g^*$ to be easy-to-learn. Identically, for the image sematic segmentation task of PSL (bottom row in Fig. 1), the event/entity ($d$) is an image lattice, the label ($g^*$) corresponding to the image lattice $d$ is a same size image lattice, in which a square represents a pixel and blue pixels indicate they belong to an object, and the target ($t^*$) is transformed from the label $g^*$ to be easy-to-learn. Here in Fig. 1, we illustrate the image classification task and the image semantic segmentation task of PSL only in two classes. From the top row in Fig. 1, we can note that, for the image classification task of PSL, the label $g^*$ can be transformed into the target $t^*$ easily, with careless designs. Since the image semantic segmentation task is the image classification task expanded to the pixel-level, for the bottom row in Fig. 1, the label $g^*$ can also be transformed into the target $t^*$ easily, with careless designs. As a result, the primary feature to distinguish PSL tasks is that the given labels $g^*$ can be directly viewed as easy-to-learn targets $t^*$ due to their precise fineness, to some extent.

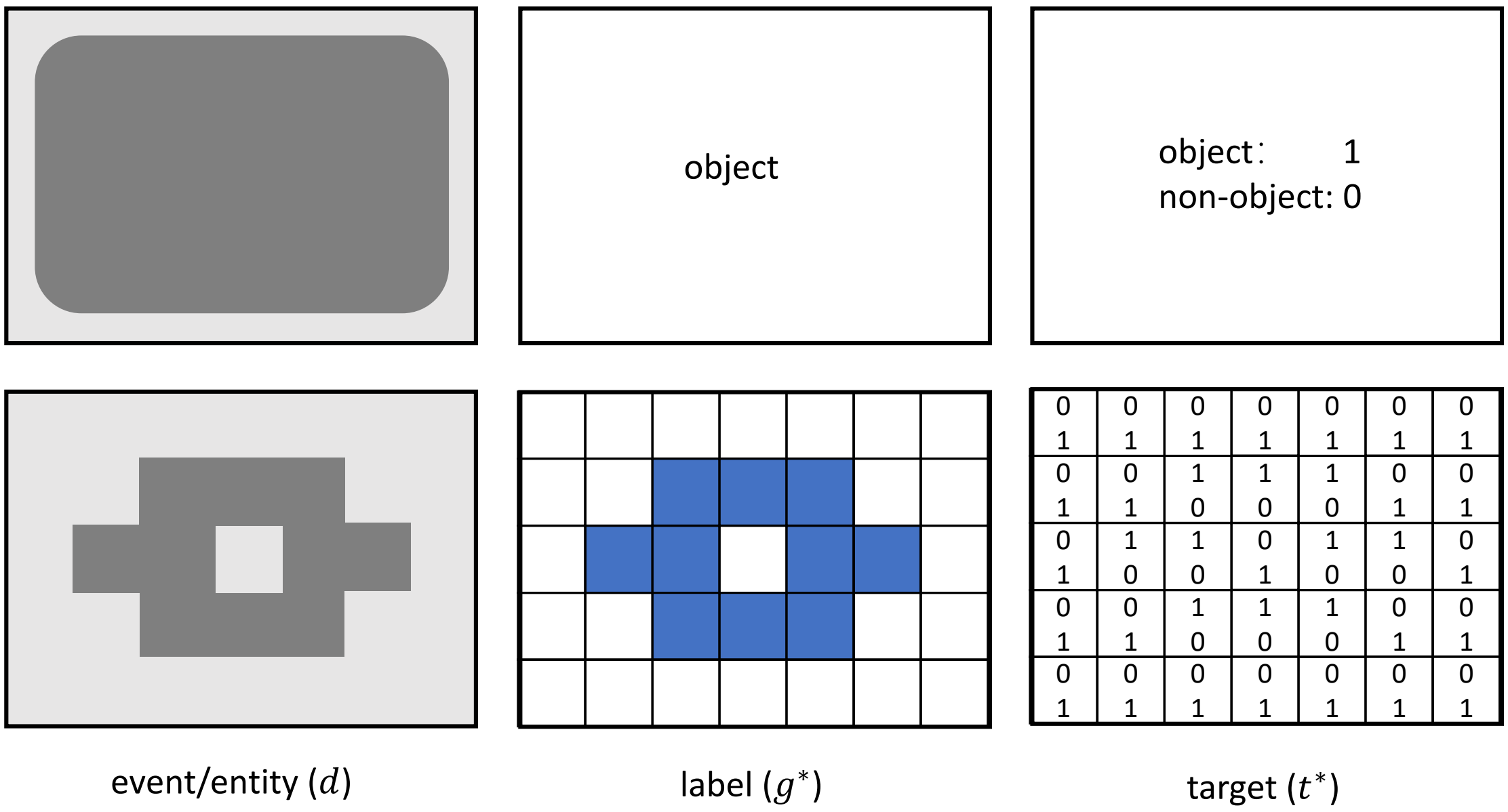

Fig. 1. Concise illustration for PSL. Top row: illustration for the typical image classification task of PSL; Bottom row: illustration for the typical image semantic segmentation task of PSL.

### 3.2.2 Moderately supervised learning

MSL focuses on the situation where the given labels $g^*$ in the training dataset are ideal while possessing to some extent extreme simplicity. This situation is different from PSL since the simplicity of $g^*$ makes directly learning with the targets from its carelessly designed transformation probably impossible or leads to very poor performance. Due to the simplicity of $g^*$, in this situation, the transformation from $g^*$ to easy-to-learn targets $t^*$ usually is parameterized and must require careful designs. Cell detection (CD) (Xie et al., 2018) and line segment detection (LSD) (Xue et al., 2019) with point labels are two typical MSL tasks.

In the CD task, the given labels for cells in an image lattice are simply a set of 2D points indicating the cell centres. In the LSD task, the given labels for line segments in an image lattice are simply a set of tuples, each of which contains two 2D points. The connection between the two 2D points of a tuple indicates a line segment in an image lattice. As the given labels for these two tasks are extremely simple, directly transforming them into easy-to-learn targets, in which pixel points corresponding to $g^*$ are set as foreground objects and the rest are set as background objects, will make the learning task impossible or lead to very poor performance. A more appropriate transformation which are restricted by a number of parameters (a parameterized transformation) can be used to alleviate this situation. However, adjusting the parameters of this parameterized transformation can result in various easy-to-learn targets that can significantly affect the performance of the final solution for an MSL task. As a result, it is usually difficult to find the optimal easy-to-learn targets from the parameterized transformation for an MSL task. Thus, an appropriately parameterized target transformation for an MSL task must require careful designs to be constructed.

In summary, a concise illustration for MSL can be shown as Fig. 2. For the cell detection task of MSL (top row in Fig. 2), the event/entity ($d$) is an image lattice, the label ($g^*$) corresponding to the image lattice $d$ is a same size image lattice, in which a coordinate indicating the centre of the cell is given, and the target ($t^*$) is transformed from the label $g^*$ to be easy-to-learn. Identically, for the line segment detection task of MSL (bottom row in Fig. 2), the event/entity ($d$) is an image lattice, the label ($g^*$) corresponding to the image lattice $d$ is a same size image lattice, in which two coordinates indicating the ends of a line segment are given, and the target ($t^*$) is transformed from the label $g^*$ to be easy-to-learn. From Fig. 2, we can note that, the labels $g^*$ for the two typical tasks of MSL are extremely simple and cannot be easily transformed into the corresponding easy-to-learn targets $t^*$, leading to the situation that the target transformations for MSL tasks requires careful designs. As a result, the primary feature to distinguish MSL tasks is that the given labels $g^*$ require careful designs to be transformed into the easy-to-learn targets $t^*$ due to their extreme simplicity.

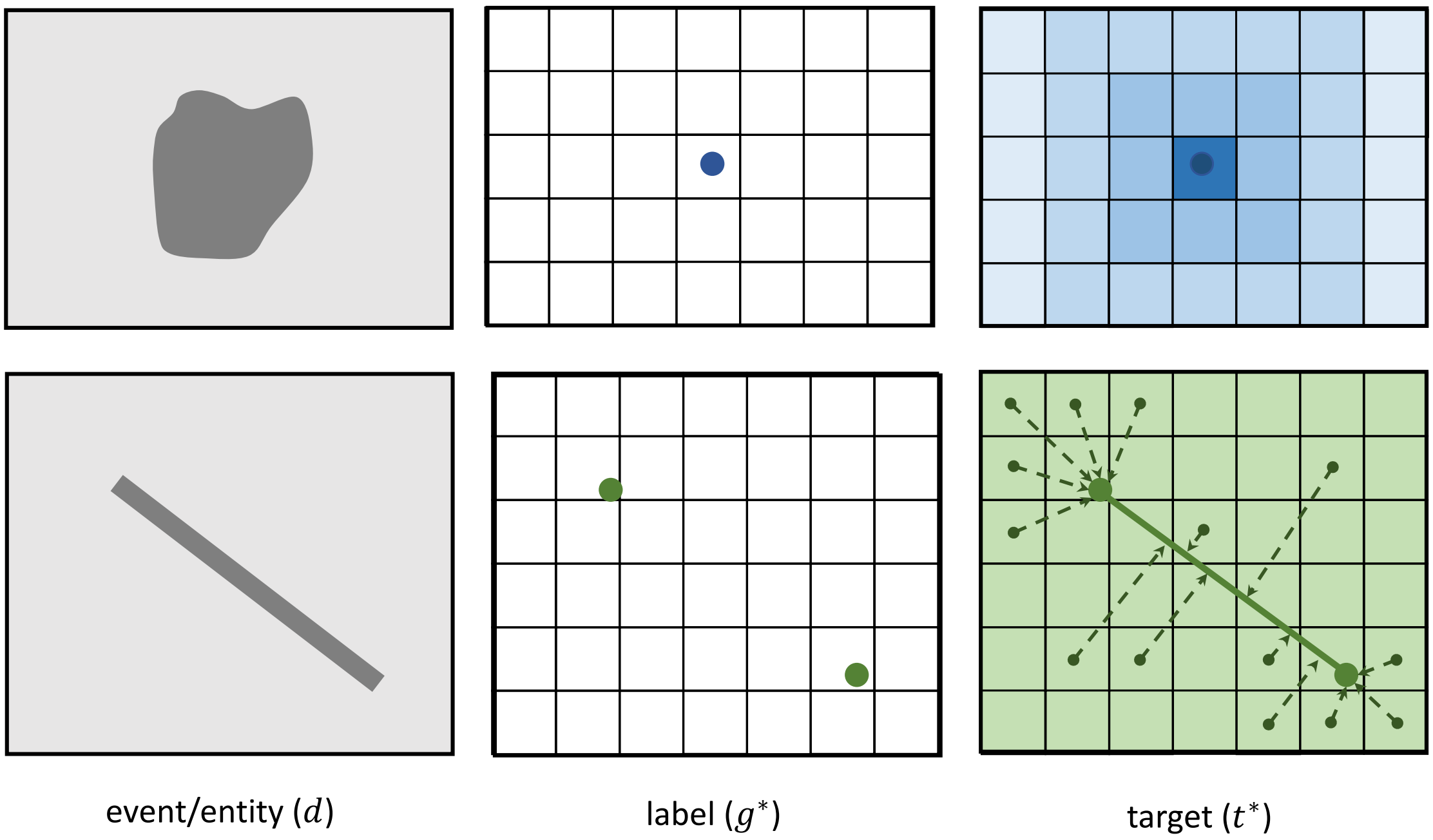

Fig. 2. Concise illustration for MSL. Top row: illustration for the typical cell detection task of MSL; Bottom row: illustration for the typical line segment detection task of MSL.

### 3.2.3 Precisely and moderately combined supervised learning

PMCSL concerns the situation where the given labels $g^*$ contain both precise and moderate annotations. In this situation, the transformation is usually built to have a mixture of properties of both the transformations designed for PSL and MSL tasks. Typical PMCSL tasks include visual object detection (Zhao et al., 2019), facial expression recognition (Ranjan et al., 2019) and human pose identification (Cao et al., 2019). Each of these tasks usually consists of a few PSL and MSL problems.

In the visual object detection task, the given labels for the objects in an image lattice are usually a set of tuples, each containing a class name and a bounding box (two coordinates) to indicate the category of an object and its position. Currently, deep convolutional neural network-based (He et al., 2016; Hu et al., 2018; Huang et al., 2017; Krizhevsky et al., 2012; Simonyan & Zisserman, 2015; Szegedy et al., 2017) one-stage approaches (YOLO (Bochkovskiy et al., 2020; Redmon et al., 2016; Redmon & Farhadi, 2017a, 2017b), SSD (Liu et al., 2016) and RetinaNet (Lin et al., 2020)) and two-stage approaches (RCNN (Girshick et al., 2014), SPPNet (He et al., 2015), Fast RCNN (Girshick, 2015), Faster RCNN (Ren et al., 2017) and FPN (Lin et al., 2017)) are the state-of-the art solutions for this task. The transformations of these solutions usually have a parameterized sub-transformation and a non-parameterized sub-transformation. The parameterized sub-transformation is responsible for pre-defining a set of reference boxes (a.k.a. anchor boxes) with different sizes and aspect ratios at different locations of an image lattice. The sizes and aspect ratios can be adjusted to generate various reference boxes. These reference boxes are used to indicate the probabilities of corresponding areas as objects in an image lattice. The non-parameterized sub-transformation is responsible for transforming the reference boxes obtained from the parameterized sub-transformation into their categories and locations according to the ground-truth class names and ground-truth bounding boxes labelled in an image lattice. Recently, researchers have also begun to propose anchor-free approaches (Duan et al., 2019; Law & Deng, 2020) to achieve object detection. In facial expression recognition (Ranjan et al., 2019) and human pose identification (Cao et al., 2019) tasks, the detection of landmarks of a face or a human is the primary problem. The given labels for the landmarks of a face or a human in an image are usually a set of tuples, each of which contains a 2D vector and a number to indicate the position and category of the landmark. The transformation of possible solution for the detection of landmarks also has a parameterized sub-transformation and a non-parameterized sub-transformation. Basically, the detection of landmarks consists of two sub-problems: locating the landmarks and classifying the located landmarks. The parameterized sub-transformation, which is similar with the target transformation for pure MSL problem, aims to generate targets for locating landmarks. And, the non-parameterized sub-transformation is responsible for producing targets for classifying the located landmarks. These typical problems show that the target transformation for PMCSL enjoys a mixture of properties of the target transformations for pure PSL and pure MSL.

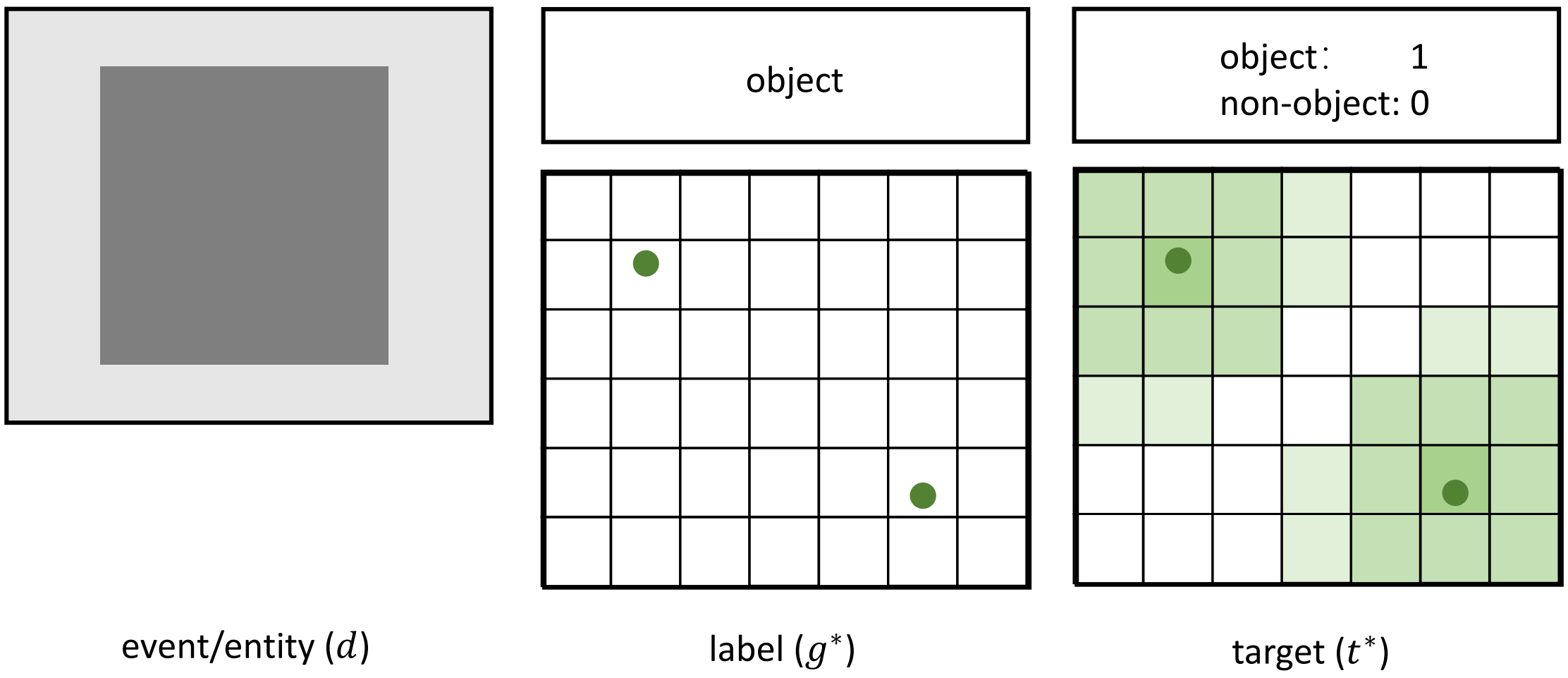

Fig. 3. Concise illustration for PMCSL.

In summary, a concise illustration for PMCSL can be shown as Fig 3. From Fig. 3, we can note that the labels $g^*$ for PMCSL tasks consist of labels for both PSL and MSL tasks, which leads to the targets $t^*$ transformed from the labels $g^*$ also consist of targets for both PSL and MSL tasks. As a result, the primary feature to distinguish MSL tasks is that the transformation from the given labels $g^*$ into the easy-to-learn targets $t^*$ constitutes of both careless and careful designs.

### 3.3 Relations of PSL, MSL and PMCSL

According to section 3.1 and section 3.2, the relations of PSL, MSL and PMCSL can be summarized as Fig. 4.

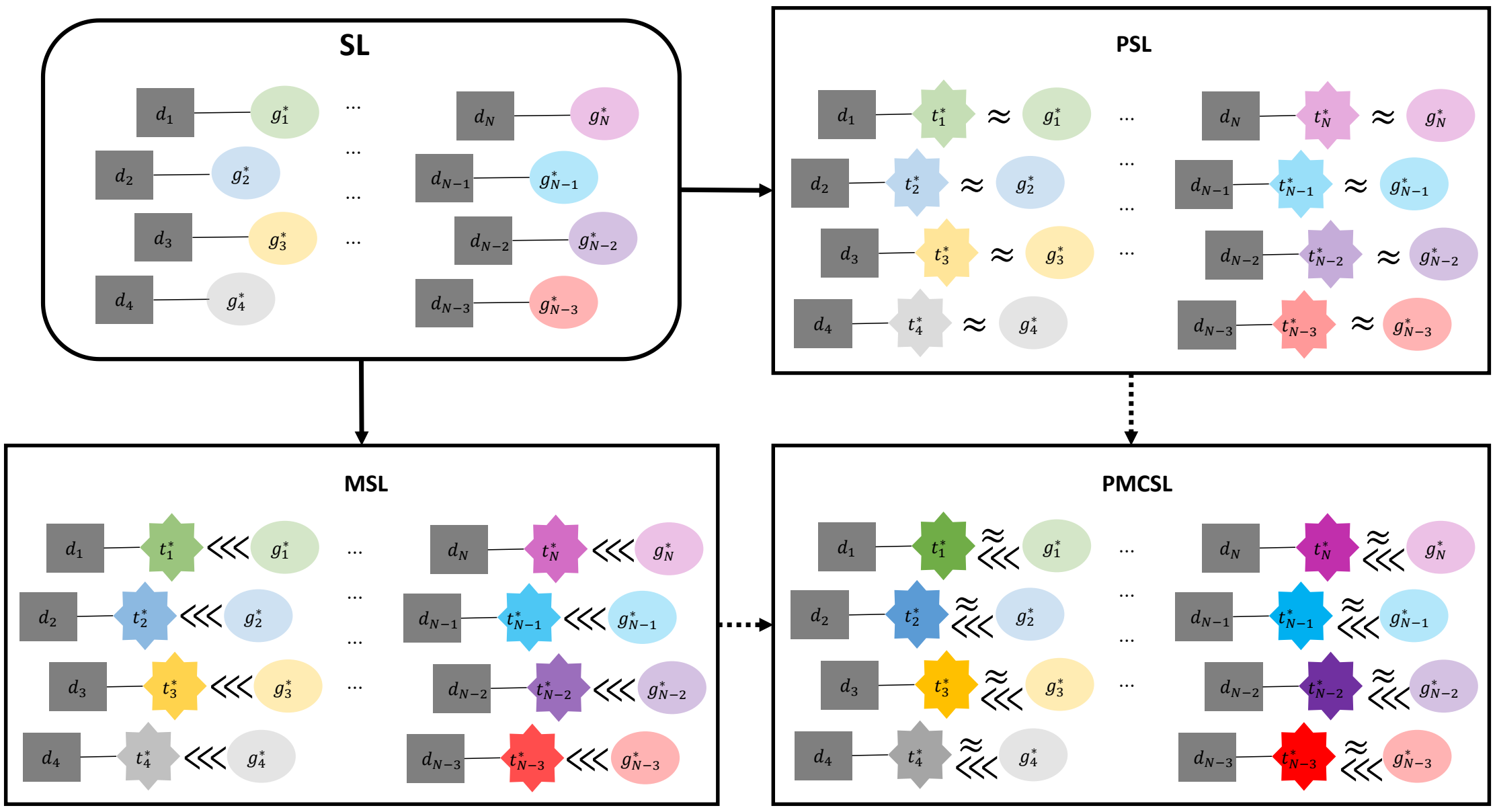

Fig. 4. Relations of PSL, MSL and PMCSL. Black rectangles denote the events/entities in the training dataset; coloured ellipses indicate the labels assigned to corresponding events/entities; coloured polygons signify the easy-to-learn targets transformed from the assigned labels. The sign $\approx$ or $\lll$ denotes the 'carelessly designed' or 'carefully designed' transformation from the assigned labels to the easy-to-learn targets.

PSL and MSL are directly derived from SL, and PMCSL is indirectly derived from SL since it is the combination of PSL and MSL. The main differences of PSL, MSL and PMCSL are their target transformations from the assigned labels to corresponding easy-to-learn targets. Usually, the target transformation of PSL is carelessly designed ($\approx$), the target transformation of MSL is carefully designed ($\lll$), and the target transformation of PMCSL constitutes both careless and careful designs ($\approx$ and $\lll$). In fact, PSL, MSL and PMCSL can be converted between each other by changing the modelling methods of the target transformations for their solutions. However, once the target transformation for a possible solution has been constructed, the sub-type of the corresponding SL task is clearly clarified. In other words, the constructed target transformation of a possible solution for an SL task fundamentally determines the sub-type of this SL task, which is crucial to building the appropriate solution for the task.

### 3.4 SL sub-types compared with WSL sub-types

WSL sub-types are determined according to the properties of the prepared labels for training data. Compared with WSL sub-types, the proposed SL sub-types are determined according to the properties of the target transformation from the given labels to easy-to-learn targets. Additionally, taking into consideration the properties presented in Definition 1 for the target transformation, the original WSL sub-types can also be classified into more refined sub-types. However, here we only focus more on the SL sub-types, since SL is more fundamental than WSL in the field of learning with supervision and the SL sub-types can naturally adjust to WSL. As a result, the proposed SL sub-types compared with WSL sub-types can be summarized as Fig. 5.

| | Prepared Labels | Target Transformation | |
|---|---|---|---|
| SL | $\mathcal{T} = \{(d_1, g_1^*), \cdots, (d_N, g_N^*)\}$ | $t^* = \{t_n^* \mid t_n^* \approx g_n^*, n \in \{1, \cdots, N\}\}$ | PSL |
| | | $t^* = \{t_n^* \mid t_n^* \lll g_n^*, n \in \{1, \cdots, N\}\}$ | MSL |
| | | $t^* = \{t_n^* \mid t_n^* \approx$ and $\lll g_n^*, n \in \{1, \cdots, N\}\}$ | PMCSL |
| Incomplete | $\mathcal{T} = \{(d_1, g_1^*), \cdots, (d_J, g_J^*), d_{J+1}, \cdots, d_N\}$ | ////////////////////////////////// | WSL |
| Inexact | $\mathcal{T} = \{(d_1, g_1^*), \cdots, (d_N, g_N^*)\}$, $d_n = \{d_{n,1}, \cdots, d_{n,M_n}\} \subseteq d$, $d_{n,m} \in d \ (m \in \{1, \cdots, M_n\})$ | | |
| Inaccurate | $\mathcal{T} = \{(d_1, (g_1^* + \Delta_1)), \cdots, (d_N, (g_N^* + \Delta_N))\}$ | | |

Fig. 5. SL sub-types compared with WSL sub-types.

# 4. Conceptualization of Moderately Supervised Learning

MSL plays the central role in the field of SL, due to the fact that PSL only counts for a minor proportion of SL and MSL is an essential part of PMCSL which account for the major proportion of SL. Although solutions have been intermittently proposed for different MSL tasks, currently, insufficient researches have been devoted to systematically analysing MSL. To fill in this gap, we present a complete fundamental basis to systematically analyse MSL, via conceptualizing MSL from the perspectives of the definition, framework and generality.

Primarily, in section 4.1, presenting the definition of MSL, we illustrate how MSL exists by viewing SL from the abstract to relatively concrete. Subsequently, in section 4.2, presenting the framework of MSL based on the definition of MSL, we illustrate how MSL tasks should be generally addressed. Finally, in section 4.3, presenting the generality of MSL based on the framework of MSL, we illustrate how generality exist among different MSL solutions by viewing different MSL solutions from the concrete to relatively abstract; that is, solutions for a wide variety of MSL tasks can be more abstractly unified into the presented framework of MSL. Particularly, the framework of MSL builds the bridge between the definition and the generality of MSL.

## 4.1 Definition

### 4.1.1 Definition of SL

Let us consider the situation where the given labels of a number of training events/entities are ideal (complete, exact and accurate) but possess simplicity. Specifically, with the given simple labels $g^*$, the ultimate goal of the learning task here is to find the final predicted labels $g$ that minimize the error against $g^*$. Regarding this situation as a classic SL problem, we can define the objective function as

$$\min_{g \in H} \ell(g, g^*), \tag{0-1}$$

where $\ell(\cdot,\cdot)$ refers to a loss function that estimates the error between two given elements. The smaller the value of this function is, the better the found $g$ is. $H$ denotes the space of the predicted targets $g$.

### 4.1.2 Definition of MSL

Due to the simplicity of $g^*$, we must carefully build a transformation that transforms $g^*$ into easy-to-learn targets $t^*$. On the basis of the transformed easy-to-learn targets $t^*$, we build a learning function that maps events/entities $d$ to the predicted targets $t$ that minimize the error against $t^*$. Based on the predicted targets $t$, we then carefully build a label re-transformation that re-transforms $t$ into the final predicted labels $g$ that can minimize the error against the labels $g^*$. We assume $t^*$ can be constructed by 'decoding' $g^*$ as the easy-to-learn targets $t^*$ are more informative than the labels $g^*$, the predicted targets $t$ can be obtained by 'inferring' $d$, and $g$ can be constructed by 'encoding' $t$ as the final predicted labels $g$ are less informative than the predicted targets $t$. Formally, we specify the following definition for MSL:

$$t^* = decoder(g^*), \quad t^* \in I,$$

$$t = inferrer(d) \ \ t \in J, \qquad \min_{t \in J} \ \ell^i(t, t^*),$$

$$g = encoder(t) \ \ g \in H, \ \min_{g \in H} \ \ell^e(g, g^*), \tag{0-2}$$

where $decoder$ denotes the target transformation, $inferrer$ denotes the learning function $f$, $encoder$ denotes the label re-transformation, and $I$ and $J$ respectively denote the space of the easy-to-learn targets $t^*$ and the space of the predicted targets $t$.

### 4.1.3 How the definition of MSL was revealed

Comparing the definition of SL (the Eq. (0-2)) with the definition of MSL (the Eq. (0-1)), we can note that, the definition of MSL was revealed by taking into consideration the transformation from the given simple labels $g^*$ to easy-to-learn targets $t^*$, which proves that some details are indeed concealed by the abstractness of the definition of SL. Intrinsically, the methodology underneath the reveal of the definition of MSL stems from viewing the SL problem from the abstract to relatively concrete, which can be summarized as Fig. 6.

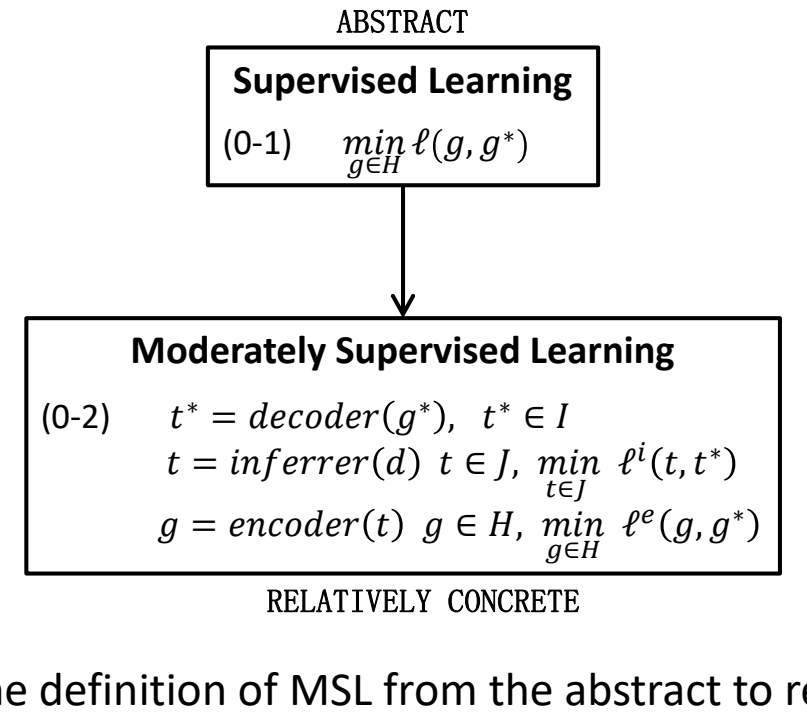

Fig. 6. Reveal of the definition of MSL from the abstract to relatively concrete.

## 4.2 Framework

On the basis of the revealed definition of MSL, in this section, we present a generalized framework for solving MSL tasks. The outline of the presented framework is shown as Fig. 7. The presented MSL framework has three basic components including decoder, inferrer and encoder, and three basic procedures including learning, looping and testing. The three basic components are the key points of constructing fundamental solutions for MSL tasks, and the learning and looping procedures are the key problems of developing better solutions for MSL tasks.

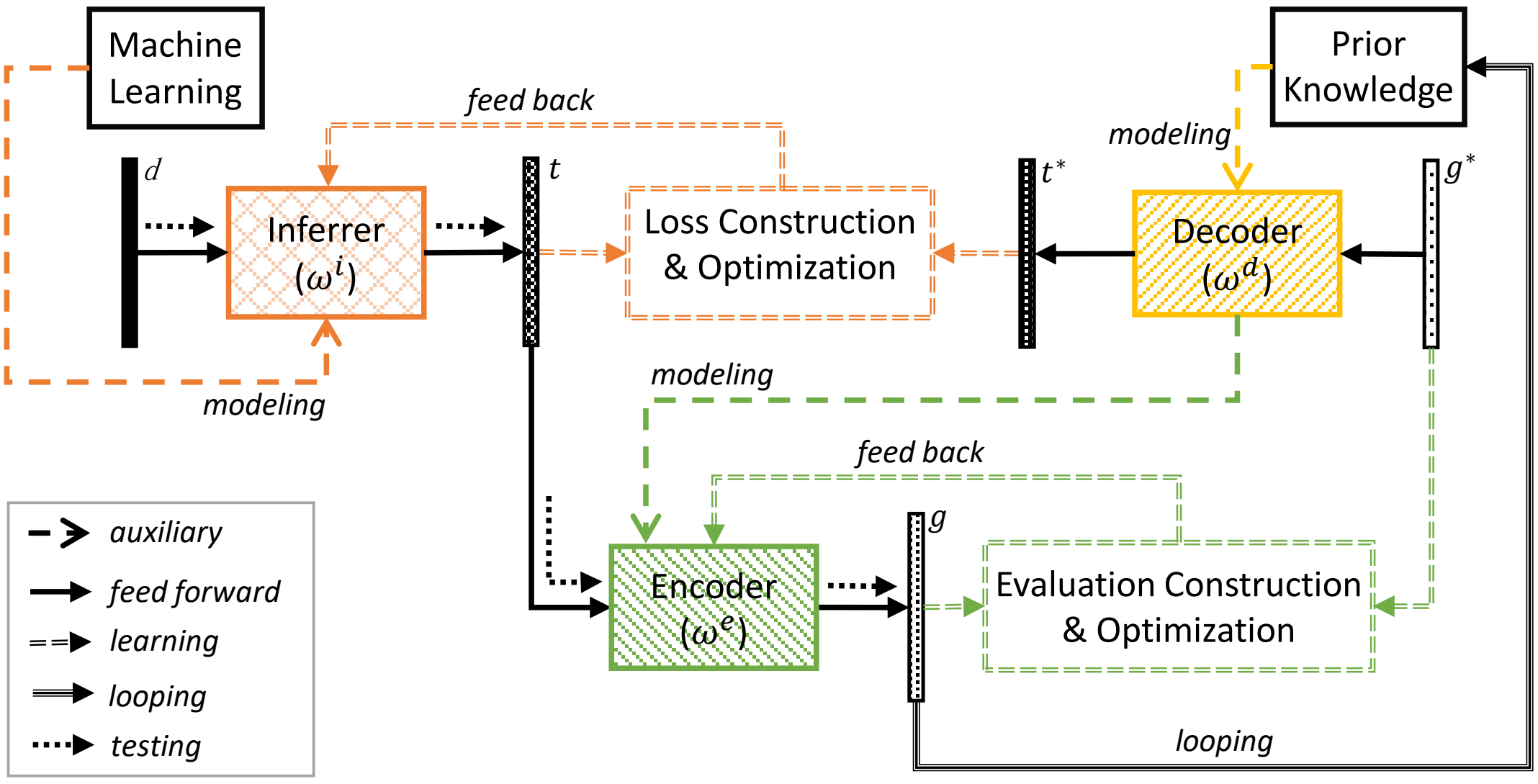

Fig. 7. Generalized framework for MSL. Basic components: decoder, inferrer and encoder. Basic procedures: learning, looping and testing.

### 4.2.1 Basic components

**Decoder** The decoder component transforms the given simple labels $g^*$ into easy-to-learn targets $t^*$. Commonly, the decoder is built on the basis of prior knowledge which is parameterized by $\omega^d$. Formally, the function of the Decoder can be expressed by

$$t^* = decoder(g^*; \omega^d). \tag{1}$$

**Inferrer** The inferrer component models the map between the events/entities $d$ and corresponding easy-to-learn targets $t^*$. Usually, the inferrer is built on the basis of machine learning techniques and is parameterized by $\omega^i$. Formally, the function of the Inferrer can be expressed by

$$t = inferrer(d; \omega^i). \tag{2}$$

**Encoder** The encoder component re-transforms the predicted targets $t$ of the inferrer into the final predicted labels $g$. Coupled with the decoder, the encoder is built on the basis of the decoder, which is parameterized by $\omega^e$. Formally, the function of the Encoder can be expressed by

$$g = encoder(t; \omega^e). \tag{3}$$

### 4.2.2 Basic procedures

**Learning** The learning procedure aims to optimize the parameters $\omega^i$ and $\omega^e$ for the inferrer and encoder, respectively, under the prerequisite of a decoder that is empirically initialized with $\overline{\omega^d}$. Specifically, we express the learning procedure as

$$\widetilde{t^*} = decoder\left(g^*; \overline{\omega^d}\right), \tag{4-1}$$

$$\widetilde{\omega^i} = arg \min_{\omega^i \in M^i} \frac{1}{N} \sum_{n=1}^{N} \ell^i\left(inferrer\left(d_n; \omega^i\right), \widetilde{t_n^*}\right), \tag{4-2}$$

$$\widetilde{\omega^e} = arg \min_{\omega^e \in M^e} \frac{1}{N} \sum_{n=1}^{N} \ell^e\left(encoder(t_n; \omega^e), g_n^*\right), \tag{4-3}$$

where $M^i$ and $M^e$ specify the parameter spaces of $\omega^i$ and $\omega^e$, respectively, and $N$ is the number of training events/entities.

**Looping** As the optimization of the parameters ($\omega^i$, $\omega^e$) for both the inferrer and encoder is conducted under the prerequisite of the decoder parameterized by $\omega^d$, a change in the decoder can significantly affect the optimization of $\omega^i$ and $\omega^e$, which will eventually be reflected in the final predicted labels $g$. In fact, prior knowledge can be enriched by analysing the predicted labels $g$ of the current solution. The enriched prior knowledge can help us to model and initialize a better decoder. Thus, in practice, we commonly loop several times to adjust the decoder and restart the training for a possibly better solution. Specifically, we express the looping procedure as

$$\widetilde{\omega^d} = arg \min_{\omega^d \in M^d} l(g|\omega^d, g^*), \tag{5}$$

where $M^d$ signifies the parameter space of $\omega^d$ and $g|\omega^d$ denotes that the final predicted labels $g$ are obtained by optimizing the parameters ($\omega^i$, $\omega^e$) of both the inferrer and encoder under the prerequisite of the decoder initialized with $\omega^d$.

**Testing** As shown in Fig. 2, testing starts from the input $d$, passes through the inferrer and encoder, and ends at $g$. Specifically, the testing procedure can be expressed as

$$t = inferrer\left(d; \widetilde{\omega^i}|\widetilde{\omega^d}\right), \tag{6-1}$$

$$g = encoder\left(t; \widetilde{\omega^e}|\widetilde{\omega^d}\right), \tag{6-2}$$

where $\widetilde{\omega^i}|\widetilde{\omega^d}$ and $\widetilde{\omega^e}|\widetilde{\omega^d}$ are the parameters of the inferrer and encoder optimized under the prerequisite of the decoder initialized with $\widetilde{\omega^d}$ found by the looping procedure.

**Framework of Moderately Supervised Learning**

**Basic components**:

(1) $t^* = decoder\left(g^*; \omega^d\right)$

(2) $t = inferrer\left(d; \omega^i\right)$

(3) $g = encoder(t; \omega^e)$

**Basic procedures**:

(4-1) $\widetilde{t^*} = decoder\left(g^*; \overline{\omega^d}\right)$

(4-2) $\widetilde{\omega^i} = arg \min_{\omega^i \in M^i} \frac{1}{N} \sum_{n=1}^{N} \ell^i\left(inferrer\left(d_n; \omega^i\right), \widetilde{t_n^*}\right)$

(4-3) $\widetilde{\omega^e} = arg \min_{\omega^e \in M^e} \frac{1}{N} \sum_{n=1}^{N} \ell^e\left(encoder(t_n; \omega^e), g_n^*\right)$

(5) $\widetilde{\omega^d} = arg \min_{\omega^d \in M^d} l\left(g|\omega^d, g^*\right)$

(6-1) $t = inferrer\left(d; \widetilde{\omega^i}|\widetilde{\omega^d}\right)$

(6-2) $g = encoder\left(t; \widetilde{\omega^e}|\widetilde{\omega^d}\right)$

Fig. 8. Formal summarization of the framework of MSL.

### 4.2.3 Analysis

From Eq. (1) to (6), the generalized framework presented for solving MSL tasks can be formally summarized as Fig. 8. We can note that Fig. 8 in fact reveals the key points of constructing fundamental solutions for MSL tasks and the key problems of developing better solutions for MSL tasks. The key points of constructing fundamental solutions for MSL tasks can be summarized as modelling the three basic components (decoder, inferrer and encoder), and the key problems of developing better solutions for MSL tasks can be summarized as evolving the learning and looping procedures to optimize the three basic components.

The decoder is responsible for transforming the given labels into easy-to-learn targets. Usually, it is built and optimized on the basis of prior knowledge. The inferrer is responsible for mapping events/entities to corresponding easy-to-learn targets. Usually, it is built and optimized on the basis of machine learning techniques. The encoder is responsible for transforming the predicted targets of the inferrer into final predicted labels. Usually, the encoder is built and optimized on the basis of the decoder.

## 4.3 Generality

Although solutions have been intermittently proposed for different MSL tasks, little work has explored the generality of different MSL tasks, due to the lack of a clear problem definition and systematic problem analysis. In this subsection, based on the specified definition and presented framework for MSL, we show that generality exists among cell detection (CD) (Xie et al., 2018) and line segment detection (LSD) (Xue et al., 2019), which are two typical MSL tasks according to the definition of MSL and have large differences in application scenarios. Following the presented framework for MSL, we review and rebuild the solutions proposed in (Xie et al., 2018; Xue et al., 2019) for these two largely different typical MSL problems to show that generality exist in their solutions.

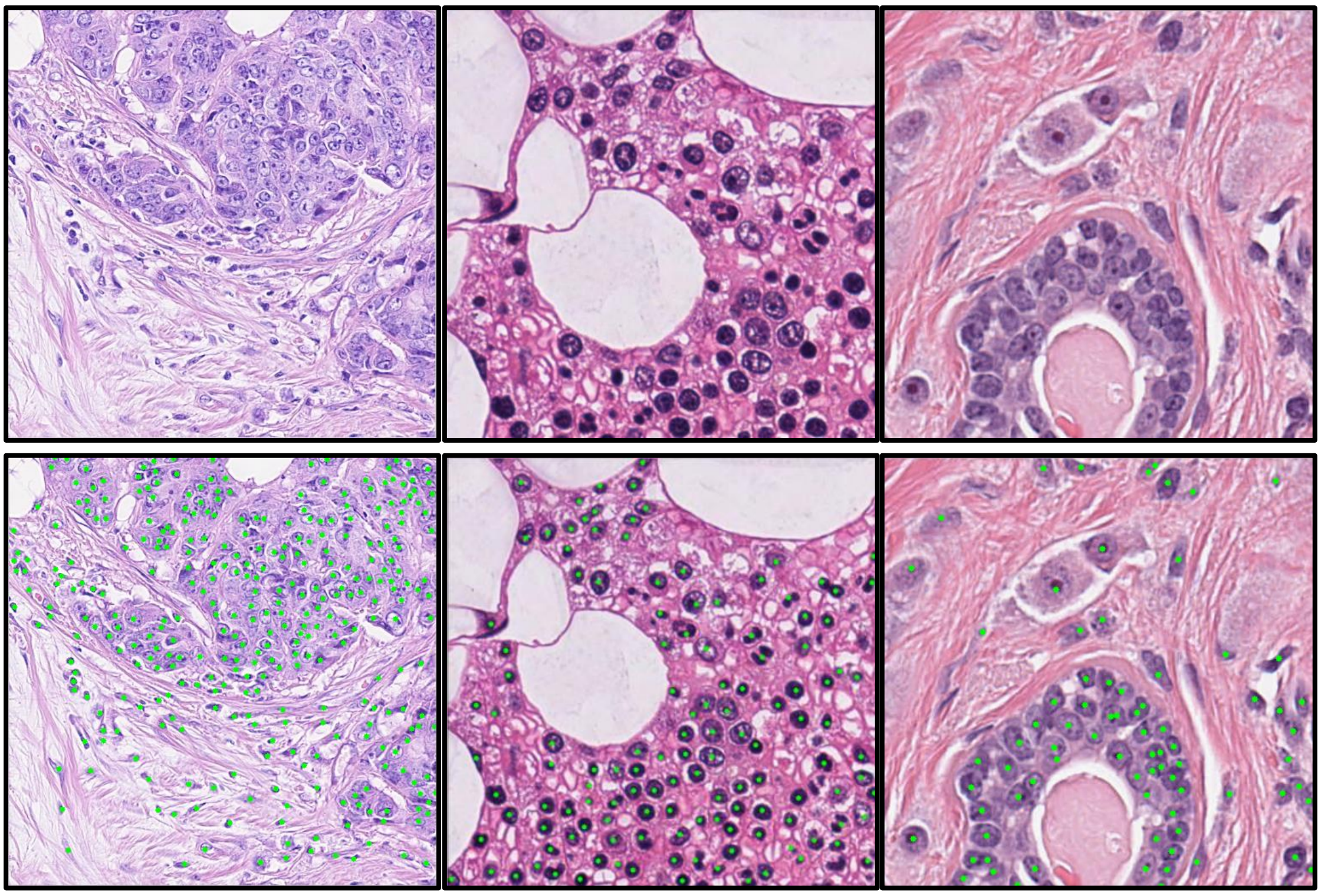

Fig. 9. Example images and corresponding labels for cell detection (Kainz et al., 2015; Sirinukunwattana et al., 2016; Xie et al., 2018). Top row: original images. Bottom row: labels shown on original images.

### 4.3.1 Cell detection

Let $\mathcal{F}$ be a 2D image lattice (e.g., 800×800). The moderate supervision information for the CD

task uses a point $p_j = (x_j, y_j)$ with offsets $x_j$ and $y_j$, respectively, to represent the cell centre in $\mathcal{F}$. In this situation, the ground-truth label in $\mathcal{F}$ is denoted by $P = \{p_j | j = \{1, \cdots, m\}\}$. Some example images and corresponding labels are given in Fig. 9.

**(1) Decoder**

The decoder transforms $P$ labelled in an image lattice $\mathcal{F}$ into a structured easy-to-learn target. It first assigns each pixel point $p \in \mathcal{F}$ to the nearest cell centre point in $P$ to partitions $\mathcal{F}$ into $m$ regions. Each region serves as a supportive area for a cell centre point. Then, by projecting its supportive region into a 1D real-valued representation, it transforms each cell centre $p_j$ in $P$ into a structured representation.

**Modelling** For assignment of a pixel point $p \in \mathcal{F}$ to its nearest cell centre point, we use Euclidian distance to define the distance between $p$ and a candidate cell centre point $p_j$ as

$$dis(p, p_j) = \sqrt{(p - p_j)^2} = \sqrt{(x - x_j)^2 + (y - y_j)^2}.$$

The supportive region for cell centre point $p_j$ is denoted by

$$R(p_j) = \{p | p \in \mathcal{F}; dis(p, p_j) < dis(p, p_k), \forall k \neq j, p_k \in P\}.$$

Then, we define the transformation of each pixel point $p$ in a supportive region $R(p_j)$ into a real-valued representation by

$$a_j(p) = e^{(1 - \frac{dis(p, p_j)}{dis_j})} - 1,\ p \in R(p_j),\ dis_j = max\, dis(p, p_j).$$

Using $a_j$, we can transform a cell centre $p_j$ into a structured 1D representation. This transformation function is applied over the entire $P$ labelled in image lattice $\mathcal{F}$ as

$$a: P \xrightarrow{transforming} \mathbb{R},$$

$$p_j \rightarrow a_j\left(R(p_j)\right), \qquad p_j \in P.$$

For simplicity, the structured target generated by the transformation function $a$ can be denoted by

$$E = \left\{a_j\left(R(p_j)\right) \middle| p_j \in P; j = \{1, \cdots, m\}\right\}.$$

**Parameterization** Using the transformation function $a$, we can transform $P$ labelled in $\mathcal{F}$ into a structured target without setting any specific parameters. However, the structured target transformed by this parameter-free model is redundant since many pixel points far from a line segment can also be assigned as being among its supportive points, which we believe is unnecessary. Note that $\varphi$ is the parameter for adjusting the selection of necessary pixel points in $\mathcal{F}$. We parameterize and rewrite the supportive region for cell centre point $p_j$ as

$$R(p_j; \varphi) = \begin{cases} R_+(p_j; \varphi) = \{p | p \in R(p_j); dis(p, p_j) < \varphi;\} \\ R_-(p_j; \varphi) = \{p | p \in R(p_j); dis(p, p_j) \geq \varphi;\} \end{cases}.$$

Then, the function for transforming each pixel point $p$ in $R(p_j; \varphi)$ into a real-valued representation is rewritten as

$$a_j(p; \{\varphi, \alpha\}) = \begin{cases} e^{\alpha(1 - \frac{dis(p, p_j)}{\varphi})} - 1 & if\ p \in R_+(p_j; \varphi) \\ 0 & if\ p \in R_-(p_j; \varphi) \end{cases},$$

where $\alpha$ is a parameter added to control the exponential decrease rate of $\varphi$ close to the centre point.

**Generalization** Decoding model: $\{R, a\}$; Input: $g^* = P$; Parameter: $\omega^d = \{\varphi, \alpha\}$; Output: $t^* = E$. Referring to Eq. (1), the transforming process of the decoder for the CD task can be defined as

$$E = \underset{\{R,a\}}{decoder}(P;\ \{\varphi, \alpha\}). \tag{CD-1}$$

Fig. 10 illustrates how the decoder transforms the ground-truth label into a easy-to-learn target for the CD task.

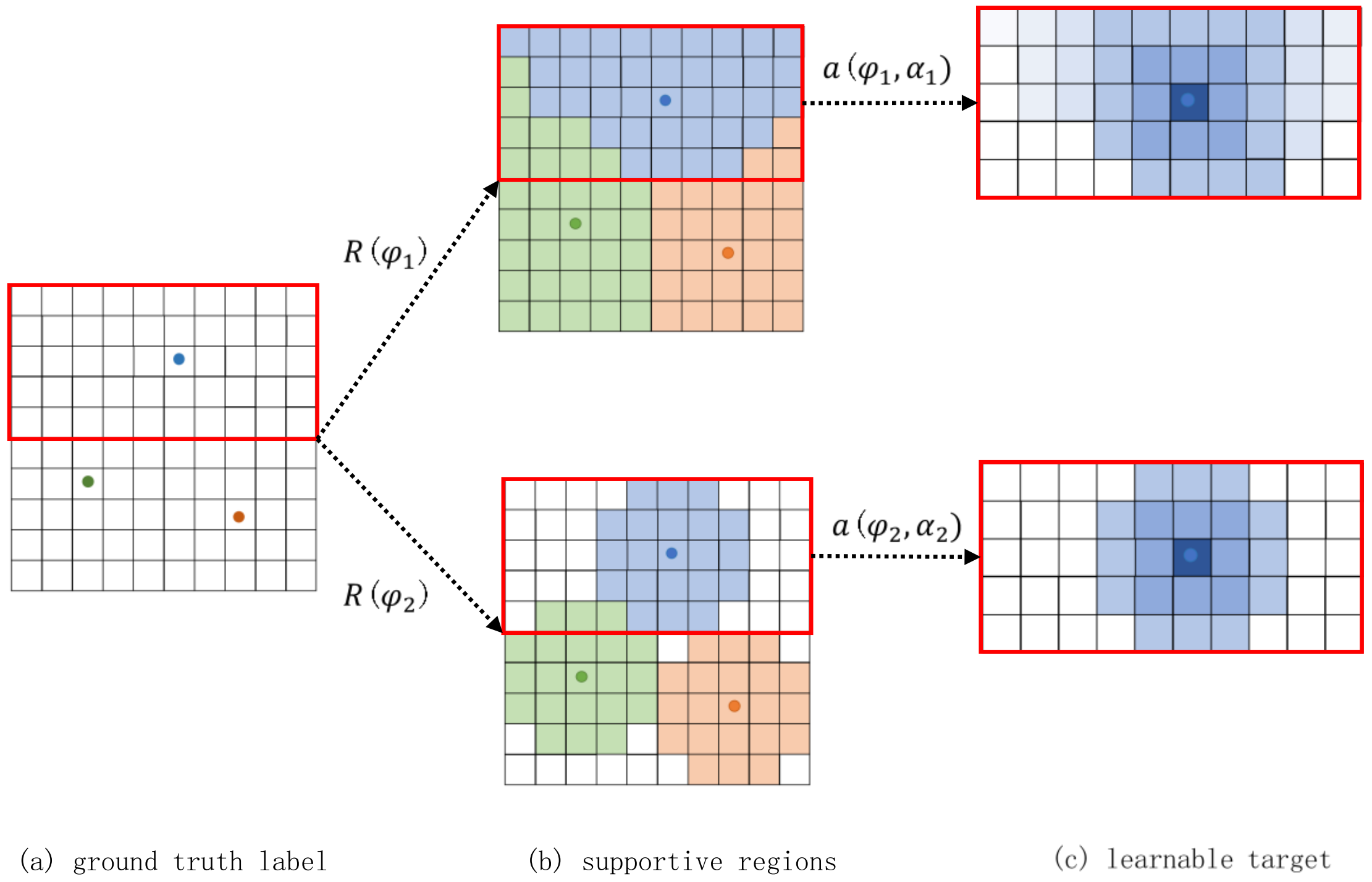

Fig. 10. Illustration of the transforming process of the decoder for the CD task. (a) The given ground-truth label is a $10 \times 10$ image lattice in which the centres of three cells are labelled. (b) Supportive regions generated during the transforming process of the decoder. (c) Easy-to-learn target generated by transforming supportive regions into a structured representation. The top rows and bottom rows of (a) and (c) are two transformations of the decoder by adjusting its parameters.

### (2) Inferrer

**Modelling** A deep convolutional neural network (DCNN) is employed to model the inferrer for mapping an input image lattice $D$ to an indirect target $G$. Define $\{f_o\}_{o=1}^{O}$ as the transformation for each of the $O$ layers from the DCNN architecture. The mapping function of the inferrer can be denoted by

$$\psi = f_O \circ f_{O-1} \circ \cdots \circ f_1.$$

Given one input image lattice $D$, the network computes the output $G$ as

$$G = \psi(D).$$

**Parameterization** We assume $\{f_o\}_{o=1}^{O}$ are parameterized by $\{\theta_o\}_{o=1}^{O}$. The corresponding $\theta_o$ has distinct forms for different types of $f_o$. The output computation of the network is rewritten by

$$G = \psi(D; \theta),$$
$$\theta = \{\theta_1, \cdots, \theta_O\}.$$

**Generalization** Inferring model: $\{\psi\}$; Input: $d = D$; Parameter: $\omega^i = \{\theta\}$; Output: $t = G$. Referring to Eq. (2), the Inferrer for the CD task can be abbreviated as

$$G = \underset{\{\psi\}}{inferrer}(D;\ \{\theta\}). \tag{CD-2}$$

**(3) Encoder**

**Modelling** Since the decoder is built to respond maximally at the cell centres, the inferrer is built to minimize the error between its output $G$ and the output $E$ of the decoder. Thus, by finding local maxima points in the lattice $G$, we can theoretically re-transform the structured prediction into the detected cell centres. For each point $p = (x, y)$ in lattice $G$, we define $p$ as a local maxima point if the values at its neighbours $N = \{(x, y-1), (x, y+1), (x-1, y), (x+1, y)\}$ are smaller than or equal to the value at point $p$, which can be expressed as

$$\phi(p) = \begin{cases} p & if\ G(p) \geq G(p_n), \forall p_n \in N \\ (-1, -1) & otherwise \end{cases}.$$

This re-transformation function is applied over the whole lattice $G$

$$\phi: G \xrightarrow{re-transforming} \mathbb{R}^2,$$

$$p \to \phi(p),\ \ p \in G.$$

As a result, the cell centres detected by the re-transformation function $\phi$ can be denoted as

$$C = \{p | p \in G; \phi(p)! = (-1, -1)\}.$$

**Parameterization** Since noise information appears in the raw output of the inferrer, we apply a small threshold $\varepsilon \in [0,1]$ to remove values smaller than $\varepsilon \cdot max(G)$. We parameterize and rewrite the re-transformation function as

$$\phi(p, \varepsilon) = \begin{cases} true & if\ G(p) \geq \varepsilon \cdot max(G) \cap G(p) \geq G(p_n), \forall p_n \in N \\ false & otherwise \end{cases}.$$

Optimizing $\varepsilon$ can balance between the recall and precision of the final cell detection.

**Generalization** Encoder model: $\{\phi\}$; Input: $t = G$; Parameter: $\omega^e = \{\varepsilon\}$; Output: $g = C$. Referring to Eq. (3), the encoder for the CD task can be abbreviated as

$$C = \underset{\{\phi\}}{encoder}(G;\ \{\varepsilon\}). \tag{CD-3}$$

**(4) Learning**

Referring to Eq. (4), in the learning procedure for the CD task, we first empirically initialize the decoder with $\{\bar{\varphi}, \bar{\alpha}\}$ and utilize the initialized decoder to transform $P$ labelled in an image lattice $\mathcal{F}$ into a structured target by

$$\bar{E} = \underset{\{R,a\}}{decoder}(P;\ \{\bar{\varphi}, \bar{\alpha}\}). \tag{CD-4-1}$$

Then, we optimize the parameters of the inferrer by

$$\tilde{\theta} = arg \min_{\theta \in M^i} \frac{1}{N} \sum_{n=1}^{N} \ell^i \left( \underset{\{\psi\}}{inferrer}(D_n;\ \{\theta\}), \bar{E} \right),$$

$$\ell^i \left( \underset{\{\psi\}}{inferrer}(D;\ \{\theta\}), \bar{E} \right) = \frac{1}{2} \sum_{p \in \mathcal{F}} (\beta G(p) + \gamma \bar{G}) \big(G(p) - \bar{E}(p)\big), \tag{CD-4-2}$$

where $\beta$ and $\gamma$ are predefined constants used to tune the losses. The parameters of the encoder are optimized by

$$\tilde{\varepsilon} = arg \min_{\varepsilon \in M^e} \frac{1}{N} \sum_{n=1}^{N} \ell^e \left( \underset{\{\phi\}}{encoder}(C_n;\ \{\varepsilon\}), P_n \right),$$

$$\ell^e\left(\underset{\{\phi\}}{encoder}(C;\ \{\varepsilon\}),P\right)=\frac{(C\cup P)-(C\cap P)}{C\cup P}. \quad \text{(CD-4-3)}$$

**(5) Looping**

Usually, we loop several times to find relatively optimal parameters for the decoder. Referring to Eq. (5), the looping procedure for the CD task can be specified as

$$\{\tilde{\varphi},\tilde{\alpha}\}=arg\ \min_{\{\varphi,\alpha\}\in M^d}\ l(C|\{\varphi,\alpha\},P),$$

$$l(C|\{\varphi,\alpha\},P)=\frac{((C|\{\varphi,\alpha\})\cup P)-((C|\{\varphi,\alpha\})\cap P)}{(C|\{\varphi,\alpha\})\cup P},$$

$$C|\{\varphi,\alpha\}=\underset{\{\phi\}}{encoder}\left(\underset{\{\psi\}}{inferrer}\left(D;\ \{\tilde{\theta}\}\right);\ \{\tilde{\varepsilon}\}\right). \quad \text{(CD-5)}$$

**(6) Testing**

Referring to Eq. (6), the testing procedure for the CD task can be specified as

$$G=\underset{\{\psi\}}{inferrer}\left(D;\ \{\tilde{\theta}\}|\{\tilde{\varphi},\tilde{\alpha}\}\right), \quad \text{(CD-6-1)}$$

$$C=\underset{\{\phi\}}{encoder}(G;\ \{\tilde{\varepsilon}\}|\{\tilde{\varphi},\tilde{\alpha}\}). \quad \text{(CD-6-2)}$$

### 4.3.2 Line segment detection

Let $\mathcal{F}$ be a 2D image lattice (e.g., 800×800). The moderate supervision information for the LSD task uses $l_k=(p_k^s,p_k^e)$ with the two points $p_k^s=(x_k^s,y_k^s)$ and $p_k^e=(x_k^e,y_k^e)$ representing the position of a line segment in $\mathcal{F}$. In this situation, the ground-truth label for $\mathcal{F}$ is denoted by $L=\{l_k|k=\{1,\cdots,m\}\}$. Some example images and corresponding labels are given in Fig. 11.

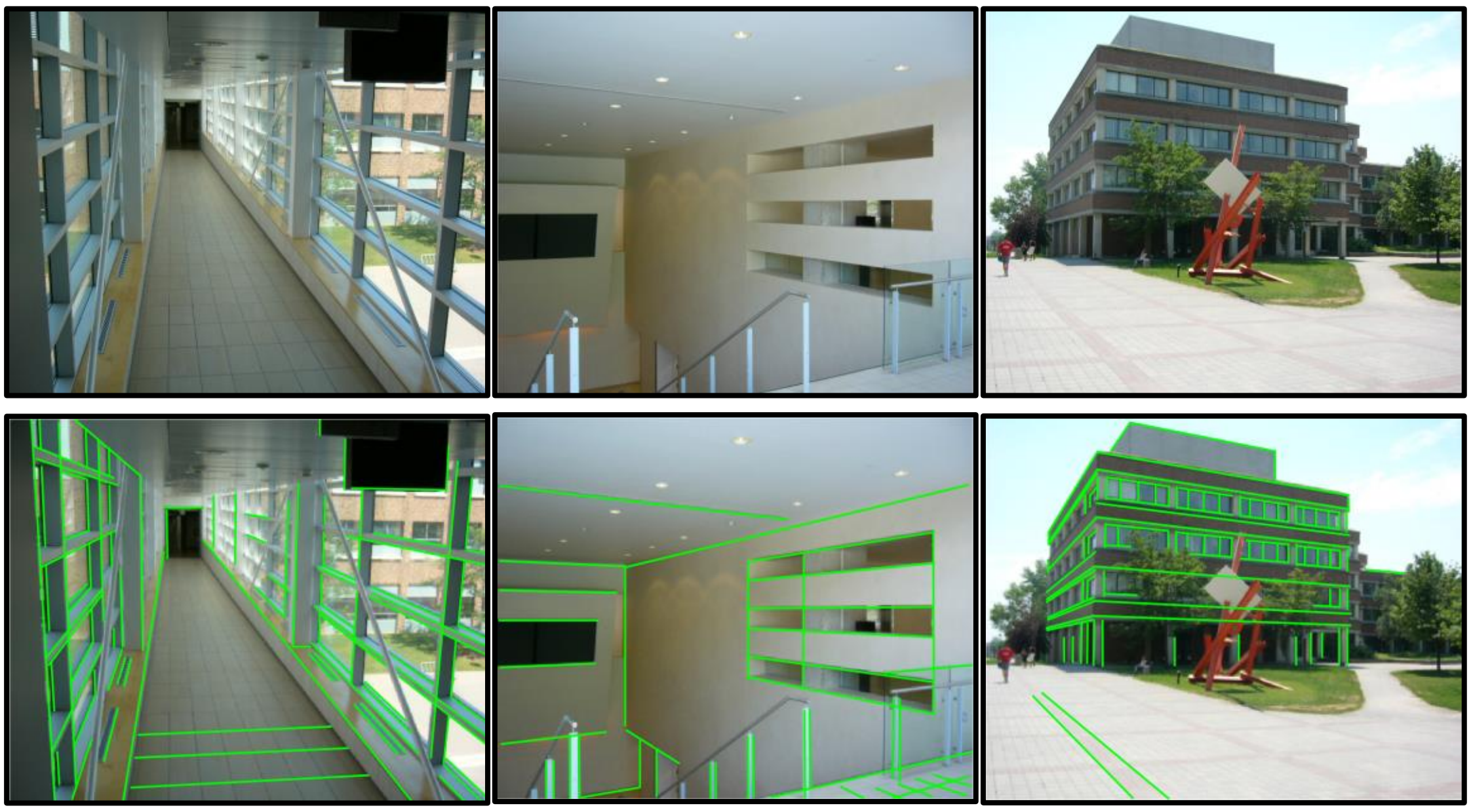

Fig. 11. Example images and corresponding labels for line segment detection (Denis et al., 2008; Matas et al., 2000). Top row: original images. Bottom row: labels shown on original images.

**(1) Decoder**

The decoder transforms line segments $L$ labelled in an image lattice $\mathcal{F}$ into a structured easy-to-learn target. It first assigns each pixel point $p\in\mathcal{F}$ to the nearest line segment in $L$, which partitions $\mathcal{F}$ into $m$ regions. Each region serves as a supportive area for a line segment. Then, by projecting its supportive region into a 2D real-valued representation, each line segment $l_j$ in $L$ is transformed into a structured representation.

**Modelling** For the assignment of a pixel point $p \in \mathcal{F}$ to its nearest line segment, we use a distance function that defines the shortest distance between $p$ and a candidate line segment $l_k$ as

$$dis(p, l_k) = \min_{z \in [0,1]} dis(p, l_k; z) = \min_{z \in [0,1]} \|p_k^s + z \cdot (p_k^e - p_k^s) - p\|_2^2.$$

The value of $t$ associated with the shortest distance between $p$ and $l_k$ is defined as

$$z_{\tilde{p}} = arg \min_{z \in [0,1]} dis\,(p, l_k; z).$$

The projection point $p'$ of $p$ on $l_k$ is defined as

$$p' = p_k^s + z_{\tilde{p}} \cdot (p_k^e - p_k^s).$$

The supportive region for line segment $l_k$ is defined as

$$R(l_k) = \{p | p \in \mathcal{F}; dis(p, l_k) < dis(p, l_j), \forall j \neq k, l_j \in L\}.$$

Then, the transformation of each pixel point $p$ in a supportive region $R(l_k)$ into a structured 2D representation is defined by

$$a_k(p) = p' - p, \qquad p \in R(l_k)$$

where the 2D representation vector is perpendicular to the line segment $l_k$ when $t_{\tilde{p}} \in (0,1)$. This transformation function is applied over the whole $L$ labelled in image lattice $\mathcal{F}$ as

$$a: L \xrightarrow{transforming} \mathbb{R}^2,$$

$$l_k \to a_k(R(l_k)), \qquad l_k \in L.$$

For simplicity, the structured target generated by the transformation function $a$ can be denoted by

$$E = \{a_k(R(l_k)) | l_k \in L; k = \{1, \cdots, m\}\}.$$

**Parameterization** Using the transformation function $a$, we can transform $L$ labelled in $\mathcal{F}$ into a structured target without setting any specific parameters. However, the structured target transformed by this parameter-free model is redundant since many pixel points far from a line segment can also be assigned as its supportive points, which we believe is unnecessary. Note that φ is the parameter for adjusting the selection of necessary pixel points in $\mathcal{F}$. We parameterize and rewrite the supportive region for line segment $l_k$ as

$$R(l_k; \varphi) = \begin{cases} R_+(l_k; \varphi) = \{p | p \in R(l_k); d(p, l_k) < \varphi\} \\ R_-(l_k; \varphi) = \{p | p \in R(l_k); d(p, l_k) \geq \varphi\} \end{cases}.$$

Then, the transformation of each pixel point $p$ in a supportive region $R(l_k; \varphi)$ into a structured 2D representation is defined as

$$a_k(p) = \begin{cases} p' - p & p \in R_+(l_k; \varphi) \\ (-x - 1, -y - 1), & p \in R_-(l_k; \varphi) \end{cases}.$$

**Generalization** decoder model: $\{R, a\}$; Input: $g^* = L$; Parameter: $\omega^d = $ φ; Output: $t^* = A$. Referring to Eq. (1), we abbreviate the transforming process of the decoder for the LSD task as

$$E = \underset{\{R,a\}}{decoder}(L;\ \varphi).$$

(LSD-1)

Fig. 12 illustrates how the decoder transforms the ground-truth label into a easy-to-learn target for the LSD task.

**(2) Inferrer**

A DCNN is employed to model the inferrer for mapping an input image lattice $D$ to an indirect target $G$, which can be denoted as

$$G = \underset{\{\psi\}}{inferrer}(D; \{\theta\}), \quad \text{(LSD-2)}$$

where $\psi$ is the mapping function of the inferrer and $\theta$ are the parameters of $\psi$.

**(3) Encoder**

**Modelling** For each pixel point $p$ in $G$, we can compute its predicted projection point on a possible line segment by

$$v(p) = p + G(p),$$

where $G(p)$ denotes the predicted 2D vector at pixel point $p$ of $G$. Its discretized point in the lattice is denoted by

$$v_{\wedge}(p) = \lfloor v(p) + 0.5 \rfloor,$$

where $\lfloor \cdot \rfloor$ represents the floor operation. In addition, if the projected point $v(p)$ is inside a line segment, $G(p)$ provides the normal direction of the line segment going through the point $v(p)$

$$\tau(p) = arctn2\left(G_x(p), G_y(p)\right).$$

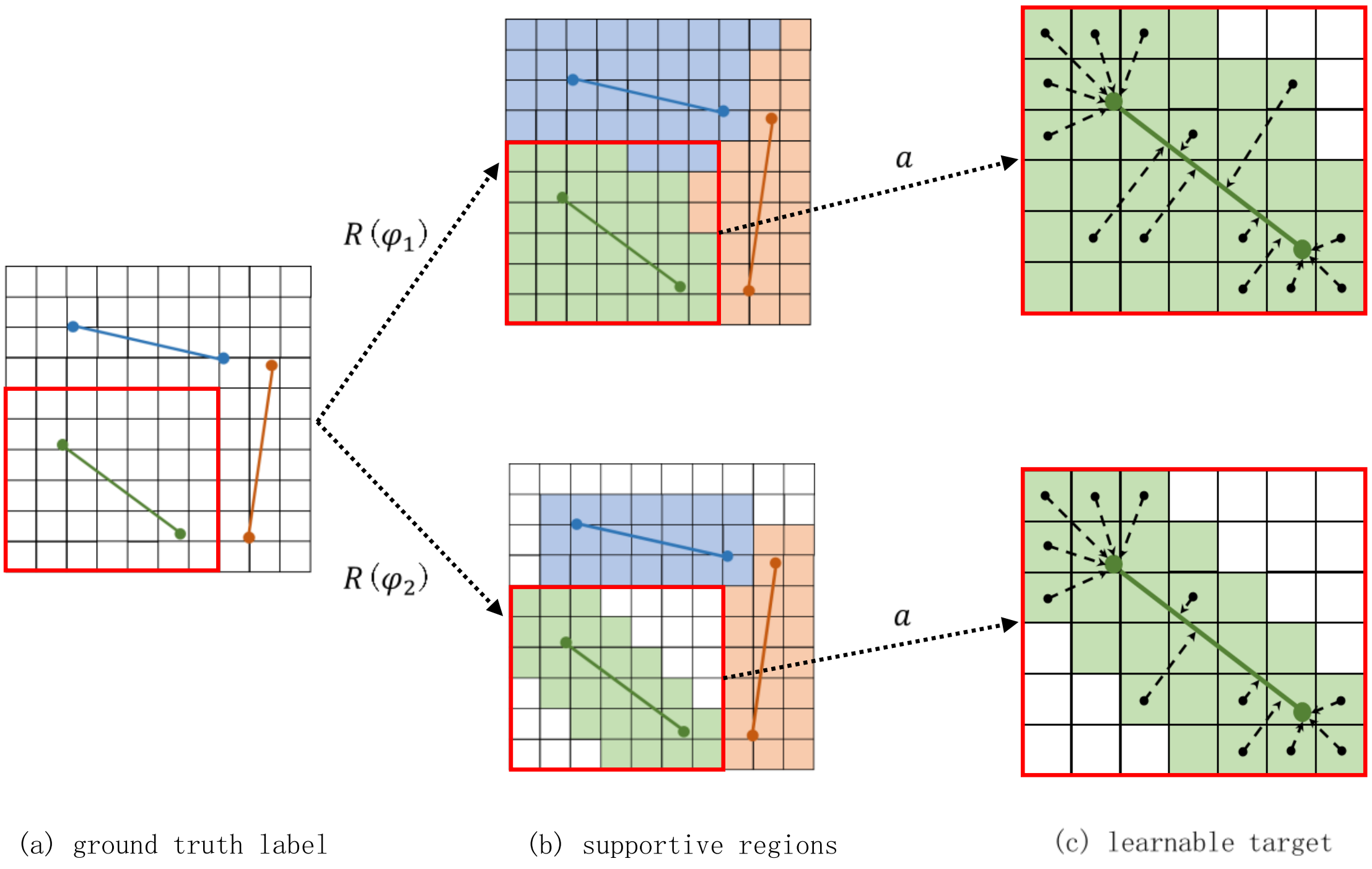

Fig. 12. Illustration of the transforming process of the decoder for the LSD task. (a) The given ground-truth label is a $10 \times 10$ image lattice in which the ends of three line-segments are labelled. (b) Supportive regions generated during the transformation process of the decoder. (c) Easy-to-learn target generated by transforming supportive regions into a structured representation. The top rows and bottom rows of (a) and (c) are two transformations of the decoder obtained by adjusting its parameters.

Using $v_{\wedge}(p)$, we can rearrange $G$ into a sparse map that records the locations of possible line segments. Such a sparse map is termed a line proposal map, in which a pixel point $q$ collects supportive pixels whose discretized projection points are $q$. The candidate set of supportive pixels collected by pixel $q$ in the line proposal map is defined by

$$S(q) = \{G(p)|p \in G, v_{\wedge}(p) = q\}.$$

Thus, using $C(q)$, a line proposal map is defined by

$$Q = \{q|S(q) \neq \emptyset, \forall q \in \mathcal{F}\}.$$

where each pixel point $q$ of $Q$ corresponds to a point on a possible line segment and is associated with a set of supportive pixels $C(q)$ in $G$. The line proposal map projects the supportive pixels for a line segment into pixels near the line segment.

With the line proposal map $Q$, the problem is to group the points of $Q$ into line segments to eventually re-transform the structured prediction into detected line segments. In the spirit of the region growing algorithm used in (von Gioi et al., 2010), an iterative and greedy grouping strategy is employed to fit line segments. The procedures of this strategy are as follows.

First, from the line proposal map $Q$, we obtain the current set of active pixels, each of which has a non-empty candidate set of supportive pixels. We also randomly select a pixel from the set of active pixels and obtain its supportive pixels. This procedure can be denoted by

$$q_0 = random_selection(Q),$$
$$c_0 = C(q_0).$$

Second, with $Q$, $q_0$, and $c_0$, we initialize the points of the current line segment and its supportive pixels and deactivate the points of the current line from $Q$, which can be denoted by

$$q_c = \{q_0\},$$
$$c_c = \{\mathrm{c}_0\},$$
$$Q = deactive(Q, q_0).$$

Iteratively, from $Q$, we find more active pixels that are aligned with the current line segment and group these active pixels aligned with points of the current line segment, which can be denoted by

$$s = center_align(Q, q_c, c_c),$$

$$group(q_c, c_c, s) => \begin{cases} end & s = \emptyset \\ \begin{cases} q_c = q_c \cup s \\ c_c = c_c \cup C(s) \\ Q = deactive(Q_a, s) \end{cases} & otherwise \end{cases}.$$

Finally, we fit the minimum outer rectangle of $q_c$ to obtain the current line segment, which can be denoted by

$$\left(p_{q_c}^s, p_{q_c}^e\right) = rectangle(q_c),$$
$$l_{q_c} = \left(p_{q_c}^s, p_{q_c}^e\right).$$

This grouping strategy is applied over $Q$ until all of its pixels are inactive. Denote the above procedures as $\phi = \{cente_align, group, rectangle\}$, the re-transformation function can be expressed as

$$\phi : G \xrightarrow{re-transforming} (\mathbb{R}^2, \mathbb{R}^2),$$

$$G \to \phi(G).$$

As a result, the line segments detected using the re-transformation function $\phi$ can be denoted by

$$C = \left\{l_{q_c} | q_c \in Q\right\}.$$

**Parameterization** We can parameterize the $center_align$ model by searching the local observation window (e.g., $r_{w \times w}$, a $w \times w$ window is used) centred at $q_c$ and finding more active pixels that are aligned with points of the current line segment with an angular distance less than a threshold (e.g., $\tau_d = 10^{\circ}$). Thus, we parameterize and rewrite $center_align$ as

$$s = center_align(Q, q_c, c_c; r_{w\times w}, \tau_d).$$

To verify the candidate line segment, we can check the aspect ratio between the width and height of the approximated rectangle with respect to a predefined threshold (e.g., $r_{w/h}$) to ensure that the approximated rectangle is sufficiently thin. As a result, we parameterize and rewrite $rectangle$ as

$$(p_{q_c}^s, p_{q_c}^e) = rectangle(q_c; r_{w/h}),$$

$$l_q = \begin{cases} \begin{cases} null \\ Q_a = reactivate(Q_a, (q_c - q_0)) \end{cases} & if\ (p_{q_c}^s, p_{q_c}^e) == (null, null) \\ (p_{q_c}^s, p_{q_c}^e) & otherwise \end{cases},$$

in which if the verification fails, we keep $q_0$ inactive and reactivate $(q_c - q_0)$.

**Generalization** Encoder model: $\phi = \{cente_align, group, rectangle\}$; Input: $t = G$; Parameter: $\omega^e = \varepsilon = \{r_{w\times w}, \tau_d, r_{w/h}\}$; Output: $g = C$. Referring to Eq. (3), the encoder for the LSD task can be abbreviated as

$$C = \underset{\{\phi\}}{encoder}(G;\ \{\varepsilon\}). \tag{LSD-3}$$

**(4) Learning**

Referring to Eq. (4), in the learning procedure for the LSD task, we can empirically initialize the decoder with $\{\bar{\varphi}\}$ and utilize the initialized decoder to transform $L$ labelled in an image lattice $\mathcal{F}$ into a structured target by

$$\bar{E} = \underset{\{R,a\}}{decoder}(L;\ \{\bar{\varphi}\}). \tag{LSD-4-1}$$

Then, we optimize the parameters of the inferrer by

$$\tilde{\theta} = arg \min_{\theta \in M^i} \frac{1}{N} \sum_{n=1}^{N} \ell^i \left( \underset{\{\psi\}}{inferrer}(D_n;\ \{\theta\}), \bar{E} \right),$$

$$\ell^i \left( \underset{\{\psi\}}{inferrer}(D;\ \{\theta\}), \bar{E} \right) = \sum_{p \in \mathcal{F}} \|\bar{E}(p) - G(p)\|_1. \tag{LSD-4-2}$$

Then, we optimize the parameters of the encoder by

$$\tilde{\varepsilon} = arg \min_{\varepsilon \in M^e} \frac{1}{N} \sum_{n=1}^{N} \ell^e \left( \underset{\{\phi\}}{encoder}(C_n;\ \{\varepsilon\}), L_n \right),$$

$$\ell^e \left( \underset{\{\phi\}}{encoder}(C;\ \{\varepsilon\}), L \right) = \frac{(C\cup L) - (C \cap L)}{C \cup L}. \tag{LSD-4-3}$$

**(5) Looping**

Usually, we loop several times to find relatively optimal parameters for the decoder. Referring to Eq. (5), the looping procedure for the LSD task can be defined as

$$\{\tilde{\varphi}\} = arg \min_{\{\varphi\} \in M^d} l(C|\{\varphi\}, L),$$

$$l(C|\{\varphi\}, L) = \frac{((C|\{\varphi\}) \cup L) - ((C|\{\varphi\}) \cap L)}{(C|\{\varphi\}) \cup L},$$

$$C|\{\varphi\} = \underset{\{\phi\}}{encoder} \left( \underset{\{\psi\}}{inferrer}(D; \{\tilde{\theta}\});\ \{\tilde{\varepsilon}\} \right). \tag{LSD-5}$$

**(6) Testing**

Referring to Eq. (6), the testing procedure for the LSD task can be defined as

$$G = \underset{\{\psi\}}{inferrer}\left(D;\ \{\tilde{\theta}\}|\{\tilde{\varphi}\}\right), \quad \text{(LSD-6-1)}$$

$$C = \underset{\{\phi\}}{encoder}(G;\ \{\tilde{\varepsilon}\}|\{\tilde{\varphi}\}). \quad \text{(LSD-6-2)}$$

### 4.3.3 How the generality of MSL was revealed

From the rebuilt solutions presented in section 4.3.1 and section 4.3.2 for the CD and LSD tasks, we can note that, though having large differences in detailed implementations, the two solutions for the CD and LSD tasks can be expressed in a similar methodological formation. In other words, by comparing the detailed implementations of the two solutions for CD and LSD from the concrete, it is hard for one to instantly say that the two solutions intrinsically share a similar methodological formation. However, by comparing the formations of the two solutions for CD and LSD from relatively abstract, it is easy for one to say that the two solutions for CD and LSD intrinsically share a similar methodological formation.

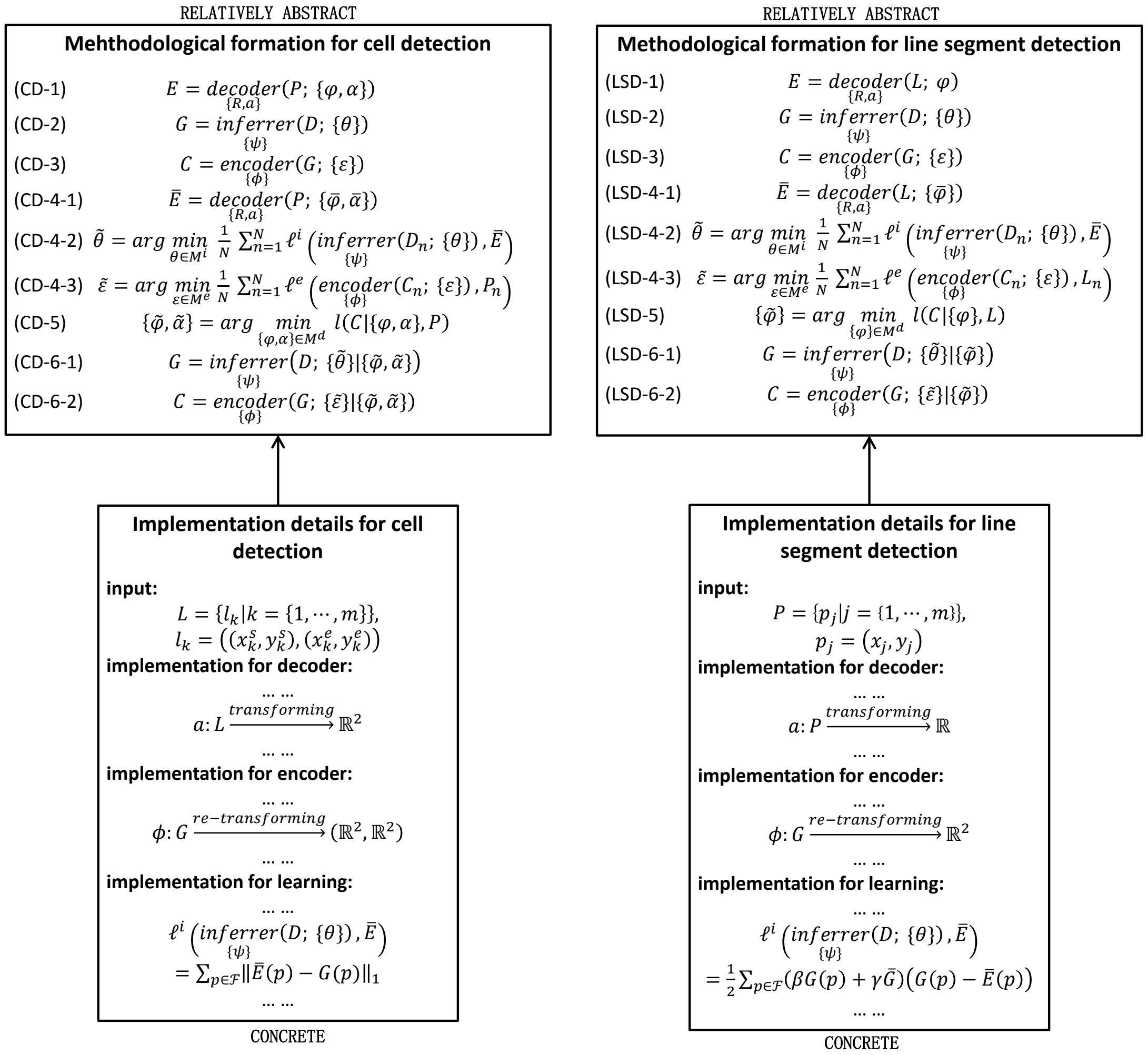

Fig. 13. Summarization of the rebuilt solutions for the CD and LSD tasks.

As a result, the generality of different MSL solutions with large differences in detailed implementations was revealed by viewing difference MSL solutions from the concrete to relatively abstract; that is, solutions for a wide variety of MSL tasks can probably be unified into a similar methodological formation. Intrinsically, the methodology underneath the reveal of the generality of MSL stems from viewing different MSL solutions from the concrete to relatively abstract, which can be summarized as Fig. 13.

## 5. Relation between conceptualization of MSL and mathematicians' vision

In this section, we reveal the relation between the conceptualization of MSL and the mathematicians' vision to illustrate the underneath methodology of proposing the new concept of MSL.

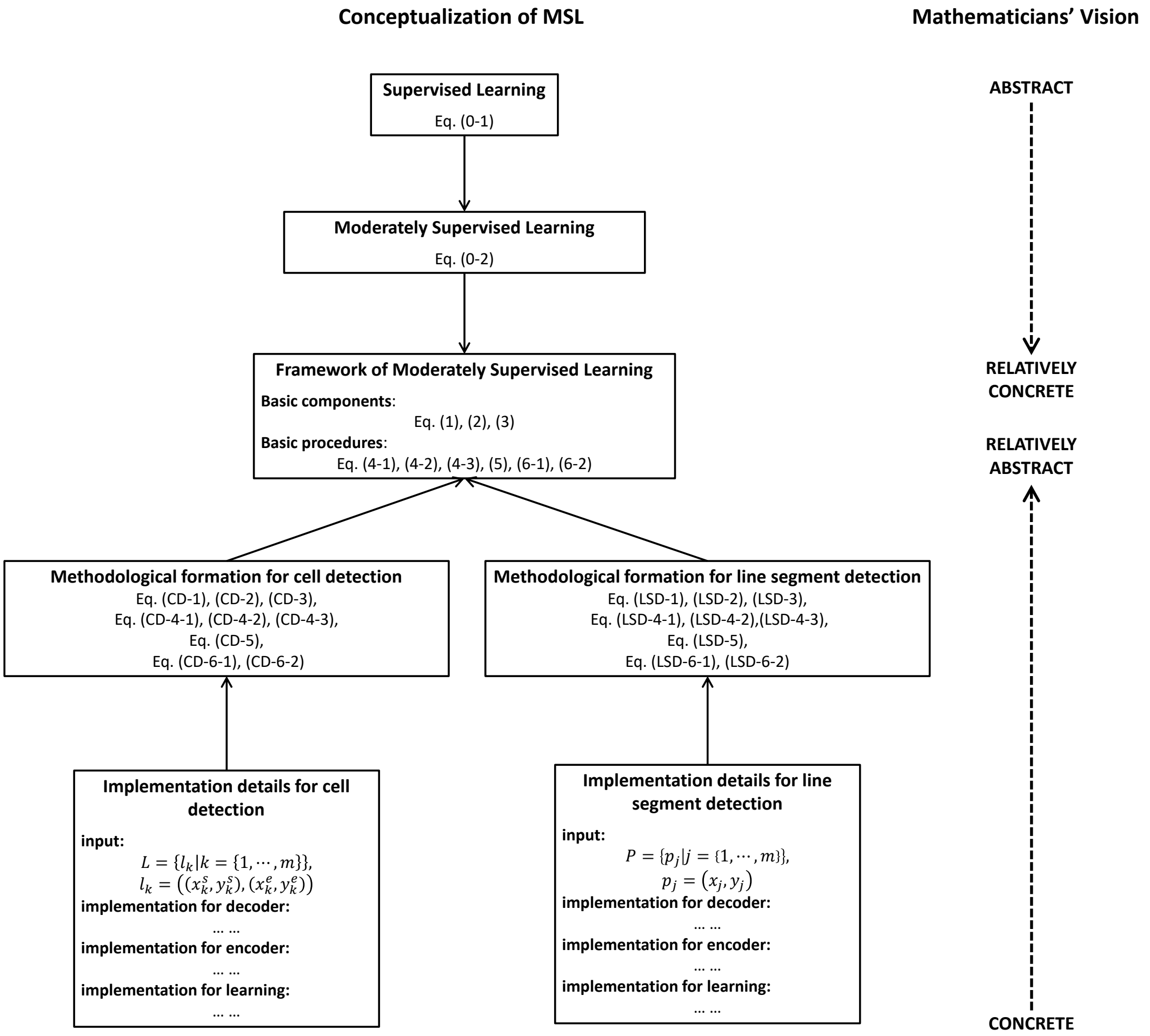

Fig. 14. Relation between the conceptualization of MSL and the mathematicians' vision.

The conceptualization of MSL presented in Section 4 has three procedures:

1) Primarily, viewing the SL problem from the abstract to relatively concrete, we reveal the existence of the definition of MSL by taking into consideration the transformation from the given simple labels to easy-to-learn targets. Specifically, viewing the SL problem from the abstract to relatively concrete is one type of mathematicians' vision, which let us notice that some important details are concealed by the abstractness of the definition of

SL.

2) Subsequently, we naturally present the framework of MSL for generally solving typical tasks based on the revealed definition of MSL. In fact, the framework of MSL builds the bridge between the definition of MSL and the generality of MSL and is an inevitable product of revealing the definition and a necessary product of revealing the generality of MSL.
3) In addition, viewing different MSL solutions from the concrete to relatively abstract, we reveal the existence of the generality of MSL by showing that MSL solutions with large differences in detailed implementations can be unified into a similar methodological formation, with natural reference to the presented framework of MSL. Specifically, viewing different MSL solutions from the concrete to relatively abstract is another type of mathematicians' vision, which let us notice that solutions for a wide variety of MSL tasks can probably share something in common even though they have large differences in detailed implementations.

As discussed in the introduction section, can we gain insight into the nature of a problem to be solved only when we look at the problem from the mathematicians' vision, which is at least both from the abstract to the concrete and from the concrete to the abstract. Specifically, without viewing the SL problem from the abstract to relatively concrete, which is one type of mathematicians' vision, the existence of the definition of MSL cannot be revealed, not to mention presenting the latter framework of MSL and revealing the generality of MSL. Similarly, without viewing different MSL solutions from the concrete to relatively abstract, which is another type of mathematicians' vision, the existence of the generality of MSL cannot be revealed. As a result, the intrinsic relation between the conceptualization of MSL presented in Section 4 and the mathematicians' vision is that the conceptualization of MSL is the product of viewing a problem to be solved from the mathematicians' vision, which can be summarized as Fig. 14.

## 6. Discussion

In the current literature, by referring to the properties of the labels prepared for the training dataset, learning with supervision is categorized as supervised learning (SL), which concerns the situation where the training dataset is assigned with ideal (complete, exact and accurate) labels, and weakly supervised learning (WSL), which concerns the situation where the training dataset is assigned with non-ideal (incomplete, inexact and inaccurate) labels. In this article, noticing the given labels are not always easy-to-learn and the transformation from the given labels to easy-to-learn targets can significantly affect the performance of the final SL solutions and taking into consideration the properties of the transformation from the given labels to easy-to-learn targets, we categorize SL into three narrower sub-types including: precisely supervised learning (PSL), which concerns the situation where the given labels are precisely fine; moderately supervised learning (MSL), which concerns the situation where the given labels are ideal, but due to the simplicity in annotation of the given labels, careful designs are required to transform the given labels into easy-to-learn targets for the learning task; and precisely and moderately combined supervised learning (PMCSL), which concerns the situation where the given labels contain both precise and moderate annotations.

Due to the fact that the MSL subtype plays the central role in the field of SL, we comprehensively conceptualize MSL from the perspectives of the definition, framework and generality. Primarily, viewing the SL problem with the mathematicians' vision from the abstract to relatively concrete, we reveal the existence of the definition of MSL by taking into consideration the transformation from the given simple labels to easy-to-learn targets. Subsequently, we naturally present the framework of MSL for generally solving typical tasks based on the revealed definition of MSL. In addition, viewing different MSL solutions with mathematicians' vision from the concrete to relatively abstract, we reveal the existence of the generality of MSL by showing that MSL solutions with large differences in detailed implementations can be unified into a similar methodological formation, with natural reference to the presented framework of MSL. The intrinsic relation between the conceptualization of MSL presented in this article and the mathematicians' vision is that the conceptualization of MSL is the product of viewing a problem to be solved from the mathematicians' vision.

As far as we know, this article is probably the first that formally and completely discussed the MSL situation in supervised learning under the era of deep learning. We hope this article will be helpful to realize that it is as well or even more critical to systematically analyse a problem to be solved, in addition to chasing the state-of-the-art result. One significance of this article is that, conceptualizing MSL from the perspectives of the definition, framework and generality, it provides the complete fundamental basis to systematically analyse the situation where the given labels are ideal, but due to the simplicity in annotation of the given labels, careful designs are required to transform the given labels into easy-to-learn targets. At meantime, the other significance of this article is that, revealing the intrinsic relation between the conceptualization of MSL and the mathematicians' vision to illustrate the underneath methodology of proposing the new concept of MSL, it establishes a tutorial of viewing a problem to be solved from the mathematicians' vision, which can be helpful for AI application practitioners to discover, evaluate and select appropriate solutions for the problem to be solved.

To end this article, we provide some possible future research directions for MSL. According to the framework of MSL, the key points of constructing fundamental MSL solutions can be summarized as modelling three basic components including decoder, inferrer and encoder, and the key problems of developing better MSL solutions can be summarized as the learning and looping procedures to optimize the three basic components. While abundant modelling approaches (Badrinarayanan et al., 2017; Chen et al., 2018; Falk et al., 2019; Shelhamer et al., 2017) and optimization methods (Bottou, 2010; Duchi et al., 2011; Kingma & Ba, 2015) have been proposed for the inferrer, the modelling and optimization of the decoder and the encoder lack systematic and comprehensive studies, except for some sporadic solutions for specific MSL tasks (Lin et al., 2017, 2020; Xie et al., 2018; Xue et al., 2019). On one hand, although successful decoders (Lin et al., 2017, 2020; Xie et al., 2018; Xue et al., 2019) have been proposed for different MSL tasks, the general methodology for modelling an appropriate decoder for an MSL task is still unclear. As the decoder determines how an MSL task is defined and is the prerequisite for optimization of both the inferrer and decoder, it is valuable to investigate how to effectively model a decoder for an MSL task with prior knowledge. On the other hand, because it is coupled with the decoder, small changes in the encoder can also significantly affect the final performance (Bodla et al., 2017; Hosang et al., 2017; Xie et al., 2018; Xue et al., 2019). Thus, it would also be interesting to investigate how to find an appropriate encoder for an MSL task. Respectively being the pre-processing and post-processing for the inferrer, the decoder and the encoder are both critical to building the appropriate solutions for MSL tasks, especially when the state-of-the-art inferrer (deep neural networks from complex (He et al., 2016; Hu et al., 2018; Huang et al., 2017; Krizhevsky et al., 2012; Simonyan & Zisserman, 2015; Szegedy et al., 2017) to lightweight (Howard et al., 2019; Sun et al., 2019; X. Zhang et al., 2018)) has been becoming standardized and reaching its limits in some specific AI applications.

## Acknowledgement

The author, Yongquan Yang, especially thanks his Yang Family, Chengdu, China for providing him with the financial supports and mental encouragements to do this research. The author really appreciates the very constructive comments received from the anonymous reviewers and editors of Information Sciences, which provided the author with the opportunities to improve the quality of this article and more confidence and support to continue to move forward with this research. The author also very much appreciates the supportive and constructive comments received from the anonymous reviewers and editors of Artificial Intelligence Review. The author thanks Professor Witold Pedrycz and Professor Feng Xia for providing the encouragements and opportunities during the publication processes of this article. Some works of this article were done in the spare time of the author while being a third-party contract employee at Institution of Clinical Pathology, West China Hospital, Sichuan University, Chengdu, China. The author thanks Professor Zhongxi Zheng and Professor Hong Bu from Institution of Clinical Pathology, West China Hospital for providing the working opportunity for a period of five years (from 2018/09 to 2023/08).